\documentclass[3p,times,procedia]{elsarticle}

\usepackage{ecrc}

\volume{00}

\firstpage{1}

\journalname{}

\runauth{}

\jid{proeng}

\jnltitlelogo{}

\CopyrightLine{2023}{Copy rights belong to the authors.}

\usepackage{amssymb}
\usepackage{amsmath}
\usepackage{bm}
\usepackage{colortbl}
\usepackage{xcolor}
\usepackage{array}
\usepackage{hyperref}
\usepackage{algorithm}
\usepackage{algorithmic}
\usepackage{booktabs}
\usepackage{tabularray}
\usepackage{multirow}

\usepackage[figuresright]{rotating}

\begin{document}

\begin{frontmatter}



\dochead{}

\title{Solving Coupled Differential Equation Groups Using PINO-CDE}


\author[a]{Wenhao Ding}
\author[a]{Qing He\corref{cor1}}
\ead{qhe@swjtu.edu.cn}
\author[b]{Hanghang Tong}
\author[a]{Qingjing Wang}
\author[a]{Ping Wang}
\cortext[cor1]{*}
\address[a]{MOE Key Laboratory of High-speed Railway Engineering, Southwest Jiaotong University, Chengdu, China}
\address[b]{Department of Computer Science, University of Illinois at Urbana-Champaign, Urbana-Champaign, United States}

\begin{abstract}
As a fundamental mathmatical tool in many engineering disciplines, coupled differential equation groups are being widely used to model complex structures containing multiple physical quantities. Engineers constantly adjust structural parameters at the design stage, which requires a highly efficient solver. The rise of deep learning technologies has offered new perspectives on this task. Unfortunately, existing black-box models suffer from poor accuracy and robustness, while the advanced methodologies of single-output operator regression cannot deal with multiple quantities simultaneously. To address these challenges, we propose PINO-CDE, a deep learning framework for solving coupled differential equation groups (CDEs) along with an equation normalization algorithm for performance enhancing. Based on the theory of physics-informed neural operator (PINO), PINO-CDE uses a single network for all quantities in a CDEs, instead of training dozens, or even hundreds of networks as in the existing literature. We demonstrate the flexibility and feasibility of PINO-CDE for one toy example and two engineering applications: vehicle-track coupled dynamics (VTCD) and reliability assessment for a four-storey building (uncertainty propagation). The performance of VTCD indicates that PINO-CDE outperforms existing software and deep learning-based methods in terms of efficiency and precision, respectively. For the uncertainty propagation task, PINO-CDE provides higher-resolution results in less than a quarter of the time incurred when using the probability density evolution method (PDEM). This framework integrates engineering dynamics and deep learning technologies and may reveal a new concept for CDEs solving and uncertainty propagation.
\end{abstract}

\begin{keyword}
Coupled differential equation group \sep Neural operator \sep Physics-informed deep learning \sep Uncertainty propagation


\end{keyword}

\end{frontmatter}


\section{Introduction}
\label{Introduction}

In the past few deacades, differential equation group has gained widespread adoption across various branches of mechanics as an essential tool for engineering mathematical modeling. The motion of each degree of freedom (DOF) in complicated systems, such as space vehicles, railway trains, and robotics, is described using a partial differential equation (PDE) or ordinary differential equation (ODE) \cite{bib1}. All the equations together form a coupled differential equation group (CDEs), the solution of which contains essential safety and performance information. With the mathematical tools derived from the increasingly sophisticated theory of numerical integration \cite{bib2}, the current challenge lies not only in solving the differential equation group, but also in optimizing the efficiency of this process.\par

Computational efficiency is critical in many engineering problems using CDEs. In a typical design project, engineers continuously adjust the structure configuration and tune the parameters prior to manufacturing. Tens of thousands of CDEs must be solved until the performance of the design meets specific requirements \cite{bib3}. A higher computing power enables more design trials prior to the deadline, resulting in better product performance. Another example is the uncertainty propagation problems associated with buildings under seismic attack \cite{bib4}, bridges under traffic loads \cite{bib5}, and offshore structures under wave excitations \cite{bib6, bib7}. In such cases, obtaining the damage probability field is the primary objective because serious casualties may occur when the structure fails. In the foreseeable future, it is unlikely that a general analytical method will be developed for large-scale and complex structures. Numerical and semi-analytical methods with statistical properties might remain mainstream, both of which require solving a large number of cases for sufficient probability resolution.\par
Unfortunately, mathematical approaches solve each equation group individually and therefore cannot learn from the solutions generated. The same applies to a physics-informed neural network \cite{bib8} (PINN). Aiming to approximate the solution function of the governing PDE with a neural network, PINN along with its variants \cite{bib9, bib10, bib11} are designed for a single PDE instead of parametric PDEs. Solving a new PDE with PINN requires retraining and often additional fine tuning \cite{bib12}, which substantially reduces its practicality in engineering CDEs. Therefore, to the best of our knowledge, all current machine-learning attempts to deal with CDEs are based on classical deep neural networks \cite{bib13, bib14}. Such black-box methods for engineering dynamics reportedly suffer from poor accuracy, interpretability, and weak generalization \cite{bib13, bib14} due to their lack of mechanics knowledge behind the data. Therefore, to fully benefit from the knowledge mechanics theory offers, we present PINO-CDE, a generalized physics-informed deep learning framework for solving engineering CDEs. The motivation for this approach stems from recent successes in deep learning for operator regression, such as ConvPDE-UQ \cite{bib15}, DeepONets \cite{bib16}, and neural operators \cite{bib17}. The concept of these approaches, and their physics-informed versions \cite{bib18, bib19, bib20}, is to learn the mapping between parameter and solution function spaces instead of approximating a single solution, such that a large number of predictions can be made with a simple forward pass once the network is trained.\par
However, the characteristics of CDEs pose additional technical challenges. The coupled differential equation group of a complicated system contains multiple physical quantities, some of which are described with PDEs (flexible objects) and others (rigid objects) with ODEs. As our goal is to present a practical framework that competes with traditional numerical integration, all quantities are considered, forming a multi-output problem. Unfortunately, although some variants consider multiple inputs \cite{bib21}, most existing operator regression approaches are designed for a single physical quantity (single output). This is particularly true for DeepONet owing to its summation structure in the last layer \cite{bib16}. To account for the multivariate situation, multiple DeepONets must be pre-trained and then concatenated in a carefully designed system \cite{bib22, bib23}. The PINO-CDE addresses this challenge by taking light from both mechanics and deep learning techniques. Instead of training dozens or even hundreds of DeepONets to map all quantities \cite{bib24}, we only need one network by using the mode superposition method \cite{bib25, bib26} (MSUP) and a fully connected neural operator structure \cite{bib27}. An equation normalization (EN) technique is also proposed to deal with the imbalance of the CDE’s equation loss scales. This technique can improve the performance of derivatives in most cases without providing their ground truth.\par
Methodologically, our framework overcomes the limitations of both classic numerical integration and machine-learning approaches for solving CDEs. Our approach offers an opportunity to increase the computing power by 3-4 orders of magnitude for solid engineering dynamics. In the following section, we first briefly introduce the workflow of PINO-CDE and elaborate on a few key techniques. Next, to demonstrate the flexibility of PINO-CDE and to validate its capability, we conducted experiments on one toy example and two practical applications: vehicle-track coupled dynamics (VTCD) and reliability analysis of a large-scale structure under seismic attack. In the following results, we show that this framework may be an important step forward in solving CDEs and probabilistic engineering.

\section{Overview of Methods and problem setup}
\label{Overview of Methods and workflow}
\begin{figure*}
\centering
\includegraphics[width=15cm]{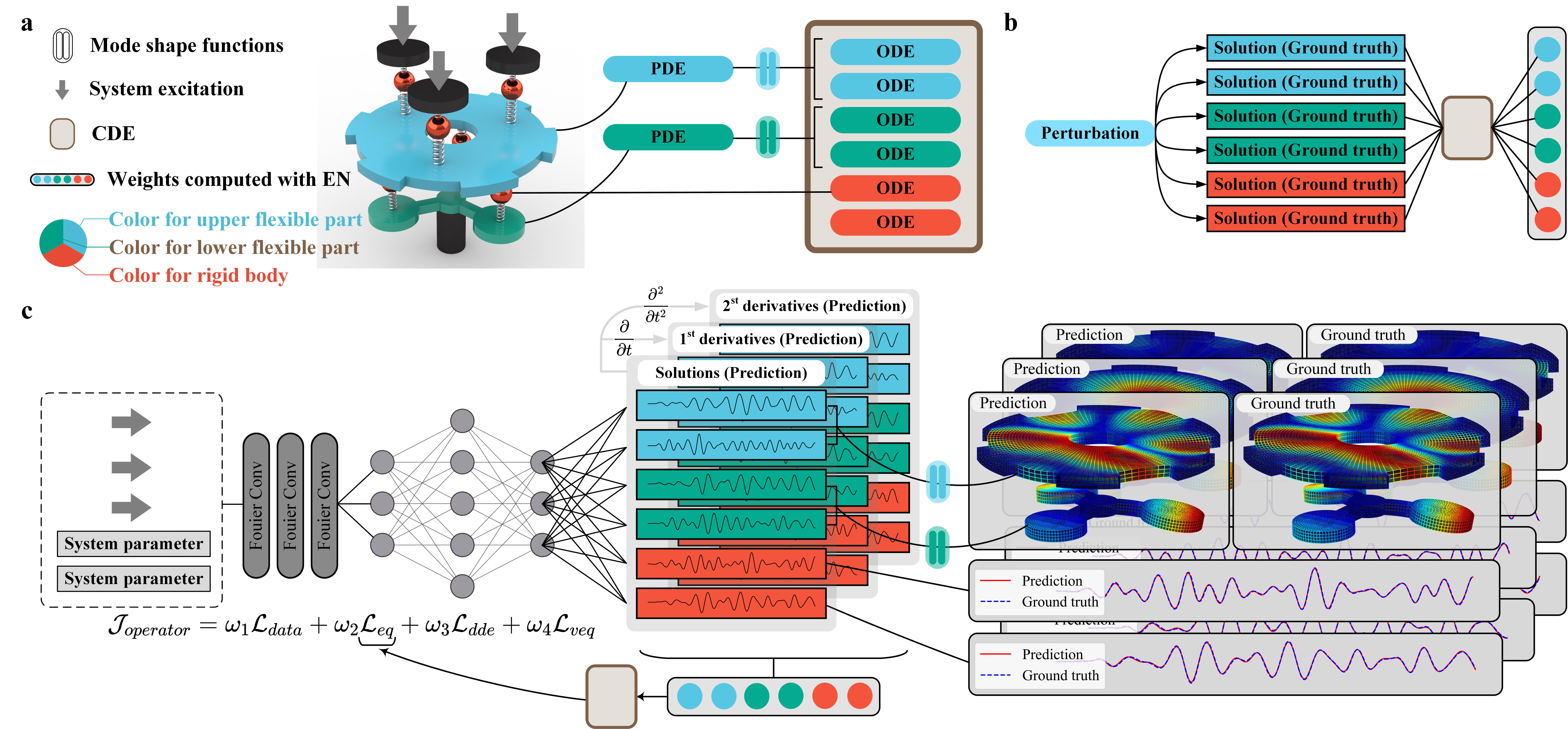}
\caption{\textbf{$\mid$ Overall workflow of the PINO-CDE. a,} Mechanical module. The governing PDE for an elastic flexible component is decomposed into multiple ODEs using the mode shape functions. Multiple components are represented by two stacked ones in the figure. All ODEs together form a coupled differential equation group for the entire system. \textbf{b,} EN technique. This technique simulates the learning process by adding perturbations into different equations, and then the weight for each equation in each data pair can be calculated accordingly. \textbf{c,} Deep learning module. The output solutions and their derivatives are fed into the physical equations with weights configured through EN to compute the equation losses, making the network physics-informed. Finally, the solution and derivative fields can be recovered with Eq. (1).}
\label{Overview_of_Methods}
\end{figure*}

In general, the PINO-CDE uses physical parameters and excitations as inputs to solve the CDEs of the targeted structure. The overall workflow can be divided into two modules: a mechanical module and a deep learning module.\par
In the mechanical module (Fig.~\ref{Overview_of_Methods}a), the PDE describing the motion of a flexible body is decomposed into multiple ODEs using the classic yet efficient MSUP \cite{bib25, bib26}. Specifically, the solution field of the flexible body is approximated by the inner product of mode shape functions and mode amplitudes as follows:
\begin{equation}
\left \{ u^\ast \right \} = \sum_{i=1}^{n} \phi_i \ast \left \{ u_i \right \} \text{,}\\
\end{equation}
where $\left \{ u^\ast \right \}$, $\phi$, $\left \{ u \right \}$, $n$, $i$ represents the solution field, mode shape functions, mode amplitudes, number of modes used, and mode index, respectively. Essentially, MSUP can be seen as a dimensional reduction operation for the data. More details are provided in Subsection \ref{Review_MSUP}.
This operation allows us to process objects with complicated boundary geometries and many details because the governing 3D PDE of elasticity is transformed into multiple simplified 1D ODEs using prior physical knowledge (mode shape functions). These ODEs, along with those describing the motions of rigid objects, form the coupled differential equation group of the system. We describe this differential equation group as follows:
\begin{equation}
\mathcal M(\bm{p},\bm{f},\bm{\mathrm{u}},t)=0,\in D \text{,}
\end{equation}
where $\bm{p}$ and $\bm{f}$ denotes the physical parameter and system excitation with $n_p$ and $n_f$ elements respectively, $\bm{\mathrm{u}}$ represents the solutions for all quantities in $\mathcal M$, and $D$ is the bounded time domain.\par
With regard to the deep learning module, a neural network is trained to approximate the solution operator for Eq. (2). Let $\bm{a}=(\bm{p,f})\in \mathcal A \subset \mathbb{R}^{n_p+n_f}$ be the variable parameter set for the mechanical system, and $\bm{\mathrm{u}}\in \mathcal U$ be the solution set with $n_{dof}$ elements. The deep learning module then aims to build an approximation $\mathcal G_{\theta}$ of the solution operator $\mathcal G^{\dagger}$ for the coupled differential equation group $\mathcal M$, which is expressed as follows:
\begin{equation}
\begin{aligned}
&\mathcal G_{\theta}: \mathcal A \to \mathcal U,\ \theta \in  \mathbb{R}^{n_p+n_f}\\
&\bm{a} \longmapsto \bm{\mathrm{u}} \\ 
\end{aligned}
\text{,}
\end{equation}
with parameter configurations from finite-dimensional space $\mathcal A \subset \mathbb{R}^{n_p+n_f}$. We used a Fourier neural operator \cite{bib27} based on graph kernel network theory \cite{bib28} as the backbone for this module. A fully connected network is connected subsequently, allowing the module to deal with multiple equations in Eq. (2) simultaneously. It is to be noted that PINO-CDE output solutions of the CDE, and their derivatives are computed through an extra numerical differential operation. The complete solution and derivative fields can then be easily recovered using Eq. (1). From the perspective of mapping approximation, PINO-CDE employs a deep neural network with FNO as its backbone to approximate the mapping between multiple 1D input signals (system physical parameters and time-varying excitations) and multiple 1D output signals (responses of respective degrees of freedom). Through MSUP, multiple 1D output signals can recover the time-varying responses of the flexible body in 3D space.\par
The loss function is constructed with the data loss $\mathcal L_{data}$, equation loss $\mathcal L_{eq}$, direct derivative loss $\mathcal L_{dde}$ and virtual equation loss $\mathcal L_{veq}$. The four components can be selected according to the application scenario, and their detailed definitions are listed in Subsection \ref{Review_PINO}. The application of $\mathcal L_{dde}$ is driven by the fact that the engineering community is generally more concerned about derivatives than solutions. For example, in nature sound is generally related to the vibration speed ($1^{st}$ derivative), while human comfort is related to acceleration ($2^{nd}$ derivative). Furthermore, we also propose an EN technique to enhance $\mathcal L_{eq}$ as an alternative because using $\mathcal L_{dde}$ leads to more GPU occupation. Details of the EN technique are provided in Subsection \ref{Review_EN}.

\section{Methodologies}

\subsection{MSUP}
\label{Review_MSUP}
The MSUP is a classic yet powerful technique for dynamics, in which the dynamic response of a structure is approximated by the superposition of a small number of its eigenmodes, as shown in Eq. (4).
\begin{equation}
\begin{aligned}
&{\left \{ u^\ast \right \} = \sum_{i=1}^{n} \phi_i \ast \left \{ u_i \right \}}  ,\\
& \left \{ u^\ast  \right \} = \text{solution field (displacement)}\\
& \left \{ \phi \right \} = \text{mode shape function}\\
& \left \{ u \right \} = \text{mode amplitude}\\
& \left \{ n \right \} = \text{number of modes used}\\
& \left \{ i \right \} = \text{mode index}
\end{aligned}
\end{equation}\par
Here, the mode shape function $\phi_{i}$ is an inherent property of the flexible object itself, which is mesh-independent and only related to the material and geometry. The analytical $\phi$ can be obtained through an analytical derivation for simple mechanical structures. For example, the rail structure in the VTCD was modeled with an Euler beam, whose analytical mode shape functions are described as follows:

\begin{equation}
Z_{k}(x)=\sqrt{\frac{2}{m_{r}l}}\sin\frac{k\pi x}{l}\text{,}
\end{equation}
where $m_{r}$ and $l$ denotes the mass per meter and total length, respectively. For more complicated objects, we used finite element method (FEM) to spatially approximate the elastic PDE and compute the numerical $\phi$. Assume that the equations for an object discretized with FEM are written in matrix form, as follows:
\begin{equation}
\bm{\mathrm{M\overset{..}{u}}}+\bm{\mathrm{C\overset{.}{u}}}+\bm{\mathrm{Ku}}=\bm{\mathrm{f}}(t)\text{,}
\end{equation}
where $\bm{\mathrm{M}}$ denotes the mass matrix, $\bm{\mathrm{C}}$ denotes the damping matrix, and $\bm{\mathrm{K}}$ denotes the stiffness matrix. The DOFs are placed in the column vector $\bm{\mathrm{u}}$ and the forces are in $\bm{\mathrm{f}}(t)$. The eigenfrequencies and corresponding mode shapes can be solved using the eigenvalue equation in Eq. (7).

\begin{equation}
(-\omega^2\bm{\mathrm{M}}+\bm{\mathrm{K}})\bm{\mathrm{u}}=0,\;u\neq0
\end{equation}

\begin{equation}
\bm{\mathrm{U}}^T\bm{\mathrm{MU\overset{..}{q}}}+\bm{\mathrm{U}}^T\bm{\mathrm{CU\overset{.}{q}}}+\bm{\mathrm{U}}^T\bm{\mathrm{KUq}}=\bm{\mathrm{U}}^T\bm{\mathrm{f}}(t)
\end{equation}

\begin{equation}
\bm{\mathrm{C}}=\alpha\bm{\mathrm{M}}+\beta\bm{\mathrm{K}}
\end{equation}

\begin{equation}
\bm{\mu\mathrm{\overset{..}{q}}}+\bm{\Omega{q}=r}(t)
\end{equation}

\begin{equation}
\gamma_{i}=\bm{\mathrm{U}}(:,i)^{T}\bm{\mathrm{M}}\bm{D},\;M_{eff}=\sum_{i=1}^{n} \gamma_{i}^2,\;M_{eff}\ge95\%
\end{equation}\par
The mode shapes (eigenmodes) are the spatial discretization of the inherent $\phi$, and only a small number $n$ of eigenfrequencies need to be computed. It is convenient to place the eigenmodes in a rectangular matrix $\bm{\mathrm{U}}$, where each column contains an eigenmode. Eq. (6) can be transformed into Eq. (8) after a left multiplication by $\bm{\mathrm{U}}^T$, and finally into Eq. (10), using the Rayleigh damping model (where $\alpha$ and $\beta$ are material-dependent constants) in Eq. (9). At this stage, the number of governing equations is reduced from the system DOF number in Eq. (6) into $n$ in Eq. (10). To retain accuracy, a sufficient $n$ must be considered to ensure a sufficient effective mass $M_{eff}$ in Eq. (11), where $\bm{D}$ is an assumed unit-displacement vector. In this study, we primarily used ABAQUS$^\circledR$ to mesh flexible objects and ANSYS$^\circledR$ to compute the mode shapes. Specifically, we used 15 and 10 modes for the upper and lower flexible components in the toy example, respectively, and 200 modes for the four-storey building in the reliability assessment, all satisfying Eq. (11).

\subsection{Physics-informed neural operator (PINO)}
\label{Review_PINO}
In the deep learning module, we used PINO \cite{bib20} with a Fourier neural operator \cite{bib27} (FNO) core as the backbone of the framework. The data features of the system excitations in the frequency domain were extracted through convolution operations in Fourier convolution layers considering different system parameter configurations. These features were then fed into the fully connected layers to be combined to output the dynamic responses for different DOFs in the targeted system. Fundamentally, the FNO aims to determine the mappings between the input and output data in the frequency domain. The convolution operations were performed with a fast Fourier transformation (FFT). The fully connected layers connected subsequently allow the deep learning module to form different feature combinations for multiple outputs. This workflow allows PINO-CDE to use a single network for all DOFs in the targeted system, thus avoiding training multiple networks as in \cite{bib22, bib23} (which is not practical for CDEs with hundreds of DOFs).\par
In the following chapters, we let PINO-CDE output solutions for 43 DOFs in the toy example (15 and 10 mode amplitudes for the upper and lower flexible objects, three rigid motion responses for each of the six marbles), 14 DOFs in the VTCD application (10 DOFs for the vehicle system and four DOFs for rail deformation under wheelsets), and 200 mode amplitudes in the reliability assessment application. Regarding the dataset size, the train and test set sizes are 1,200 and 1,000 for the toy example, 10,000 and 2,000 for VTCD, and 150 and 25 for the reliability analysis.\par
During training, the PINO-CDE harvests gradients from four types of loss functions: data loss $\mathcal L_{data}$, equation loss $\mathcal L_{eq}$, direct derivative loss $\mathcal L_{dde}$, and virtual equation loss $\mathcal L_{veq}$.

\begin{equation}
\begin{gathered}
\mathcal J_{operator}=\omega_1\mathcal L_{data} + \omega_2\mathcal L_{eq} + \omega_3\mathcal L_{dde} + \omega_4\mathcal L_{veq}\\
\mathcal L_{data}(\bm{u}, \mathcal G_\theta(\bm(a))=\mathbb E_{\bm{a}\sim \bm{u}}\left [ \left \| \bm{u}-\mathcal G_{\theta}(\bm{a}) \right \|_{\mathcal{U}}^{2} \right ]\approx \frac{1}{N} \sum_{j=1}^{N} \int_{T}^{} \mid {u(t)-\mathcal G_{\theta}(a)(t)} \mid^2 dt\\
\mathcal L_{eq}=\mathbb E_{\bm{a}\sim \bm{u}}\left [ \left \| \mathcal M(\bm{a},\mathcal G_{\theta}(\bm{a})) \right \|_{\mathcal{U}}^{2} \right ] \approx \frac{1}{N} \sum_{j=1}^{N} \int_{T}^{} \mid \mathcal M(\bm{a},\mathcal G_{\theta}(\bm{a}),t) \mid^2 dt\\
\mathcal L_{dde}=\mathbb E_{\bm{a}\sim \bm{u}}\left [ \left \| \left ({\frac{\partial \mathcal \bm{u}}{\partial t} + \frac{\partial^2 \mathcal \bm{u}}{\partial t^2}}\right )-\left ({\frac{\partial \mathcal G_{\theta}(\bm{a})}{\partial t} + \frac{\partial^2 \mathcal G_{\theta}(\bm{a})}{\partial t^2}}\right )  \right \|_{\mathcal U}^{2} \right ] \\
\approx \frac{1}{N} \sum_{j=1}^{N} \int_{T}^{} \mid \left ({\frac{\partial \mathcal \bm{u}}{\partial t} + \frac{\partial^2 \mathcal \bm{u}}{\partial t^2}}\right )-\left ({\frac{\partial \mathcal G_{\theta}(\bm{a})}{\partial t} + \frac{\partial^2 \mathcal G_{\theta}(\bm{a})}{\partial t^2}}\right ) \mid^2 dt
\end{gathered}
\end{equation}\par

We assume a dataset $\left \{ \bm{a}_j,\bm{u}_j \right \}_{j=1}^{N}$ is available, where $\bm{a}_j \sim \bm{\mu}$ are parameter configurations sampled from some multi-dimensional distribution $\bm{\mu}$ supported on $\mathcal A$. Before training, data normalization is performed for $\bm{a}_j$ and $\bm{u}_j$, ensuring that data losses $\mathcal L_{data}$ obtained from different DOFs are comparable. Equation loss $\mathcal L_{eq}$ in Eq. (12) is defined as the average squared norm value over the time domain in $\mathcal A$. During the training process, PINO-CDE outputs only the solutions for the targeted system, and their derivatives are computed through a differential operation after being renormalized. We used a simple numerical difference operation in the time domain to achieve this, considering its comprehensive advantages of efficiency, low GPU occupancy and stability. These data were then fed into the differential equation group $\mathcal M$ to generate the equation loss $\mathcal L_{eq}$. $\mathcal L_{veq}$ shares the same expression as $\mathcal L_{eq}$, except that it is sampled from an additional dataset $\left \{ \bm{a}_{k} \right \}_{k=1}^{M}$, where the parameter configurations are provided without the corresponding ground truth for the solutions. In the case of small datasets, $\mathcal L_{veq}$ enables PINO-CDE to obtain equation losses on a larger dataset. This hardly increases the training cost since no corresponding ground truth needs to be generated.\par

Finally, $\mathcal L_{dde}$ is computed by directly providing PINO-CDE with the ground truth of the $1^{st}$ and $2^{nd}$ derivatives. Undoubtedly, using $\mathcal L_{dde}$ positively affects the accuracy of derivatives by taking advantage of the ground truth data for derivatives, which is generally wasted in existing research. This is especially true for derivative-insensitive equations, where the accuracy of derivatives has little effect on their equation losses. However, although generating the ground truth data for derivatives does not increase the cost significantly (most numerical integration algorithms can provide both the solutions and their derivatives at the same time), using $\mathcal L_{dde}$ requires more GPU occupation, thus slowing down the training process. Furthermore, for some cases (denoted by $+$ in Table~\ref{tab1}), we only computed $\mathcal L_{dde}$ for a small period of time on the edges of the time domain. This will help PINO-CDE  impose sufficient boundary constraints on the time domain to improve the performance without significantly increasing the training cost. The four loss functions were balanced with weights $\omega_i,(i=1\sim 4)$ controlled by the GradNorm technique \cite{bib46}, and different combinations can be selected for different purposes. For example, users looking for precision may tend to use $\mathcal L_{dde}$, whereas those seeking to reduce training costs may tend to use $\mathcal L_{eq}$.\\

\subsection{The EN technique}
\label{Review_EN}
For common operator regression problems with a single PDE, applying a manually determined weight for each type of loss function is sufficient. However, this does not apply to CDEs because the differential equation group $\mathcal M$ contains $n_{dof}$ ODEs, each describing a unique physical motion. During the training process, the scales of the equation losses for different ODEs in $\mathcal M$ can vary significantly when their data losses are similar. Because of this CDEs-unique imbalance phenomenon, most equation losses harvested by PINO-CDE are only from sensitive ODEs, that is, ODEs in which the accuracy of the solutions has a significant influence on their equation losses. However, the sensitivity of an equation is merely a mathematical feature and cannot represent the importance of the physical quantity it describes. To address this, we propose an EN technique that fully utilizes prior physics knowledge.\par
Specifically, EN generates a weight $\lambda$ for each ODE in each data pair, which can be used to normalize their equation losses during training. These weights allow the PINO-CDE to approximate different ODEs uniformly during training, thus improving the overall performance. The full EN algorithm is summarized in Algorithm~\ref{algo1}.\par
Here, $\sigma \left ( x  \right )$ represents the standard deviation of $x$, and $\varepsilon \left ( a,b \right ) $ generates a random sequence following a uniform distribution on $\left ( -a,a \right ) $, with the same dimension as $b$. $\max^T \mid \mathcal M \mid$ computes the extreme value in the time domain for each ODE in $\mathcal M$. $r$ is a predefined constant representing the acceptable error level, which is set as $2\%$ in this study. During training, $\lambda_{ij}$ is allocated to each ODE in each data pair, which enables their equation losses to be normalized to approximate $r$ when their data losses are close to $r$. From another perspective, adding perturbation to the ground truth and computing the corresponding equation losses is, in fact, a simulation of the learning process. The degree of difference in the equation losses for different ODEs indicates the severity of the imbalance (see Fig.~\ref{Case1_Recovered_Performance}b, Fig.~\ref{Case2_Analysis_on_PINO-CDE_Performance_for_VTCD}a in following chapters). 
\begin{algorithm}
\caption{Compute weight $\lambda_{ij}$ for the $j^{th}$ ODE in the $i^{th}$ data pair}\label{algo1}
\begin{algorithmic}[1]
\FOR{$i = 1 \to N$}
	\STATE Extract data pair $\left \{ \bm{a}_i,\bm{u}_i \right \}$
	\FOR{$j = 1 \to n_{dof}$}
		\STATE $u=\bm{u}_i\left ( :,j \right ) $
		\STATE $p_0 \Leftarrow \varepsilon \left ( r*\sigma \left ( u \right ),u \right ) , p_0 \Leftarrow \varepsilon \left ( r*\sigma \left ( \dot{u}  \right ),\dot{u}  \right ), p_0 \Leftarrow \varepsilon \left ( r*\sigma \left ( \ddot{u}  \right ),\ddot{u}  \right ) $
		\STATE $\bm{s_0}\left ( :,j \right )  \Leftarrow u+p_0, \bm{s_1}\left ( :,j \right )  \Leftarrow \frac{\partial u}{\partial t}+p_1, \bm{s_2}\left ( :,j \right )  \Leftarrow \frac{\partial^2 u}{\partial t^2}+p_2, $
	\ENDFOR
	\STATE $\bm{L}=\max^T \mid \mathcal M(\bm{a_i},\bm{s_0},\bm{s_1},\bm{s_2}) \mid$
	\FOR{$j = 1 \to n_{dof}$}
		\STATE $\lambda_{ij} =\frac{r}{{\bm{L}}\left ( j \right ) }$
	\ENDFOR
\ENDFOR
\end{algorithmic}
\end{algorithm}

\subsection{PDEM}
\label{Review_PDEM}
The PDEM was developed by Li and Chen at the beginning of this century based on the principle of probability preservation \cite{bib36, bib37, bib38}. The past two decades have witnessed its successful applications on many stochastic engineering problems, particularly seismic analyses. Therefore, we use the PDEM in this work to perform a reliability analysis for a four-storey building as a baseline for PINO-CDE to compete with. The governing equation of the PDEM is written as:
\begin{equation}
\frac{\partial {p_{X_{\bm a}}\left ( x,\bm{a},t \right ) }}{\partial t} +\dot{X} \left ( \bm{a},t \right )\frac{\partial {p_{X_{\bm a}}\left ( x,\bm{a},t \right ) }}{\partial x}=0
\end{equation}
and the probability density function of $X \left ( t \right) $ reads
\begin{equation}
p_{X}\left ( x,t \right ) =\int\limits_{\mathcal A}^{} p_{X_{\bm a}}\left ( x,\bm{a},t \right ) \mathrm{d}\bm{a} \text{,}
\end{equation}
where $X$ denotes the target physical quantity; $\bm{a}$ is the random parameters involved; $p_{X_{\bm a}}\left ( x,\bm{a},t \right ) $ is the joint probability density function of $\left ( X,\bm{a} \right ) $; $\mathcal A$ is the random parameters space; $p_{X}\left ( x,t \right )$ represents the PDF of $X\left ( t \right ) $.\par
Eq. (13) is a 1D convection PDE bridging the joint PDF of $\left ( X,\bm{a} \right ) $ and the corresponding velocity $\dot{X} \left ( \bm{a},t \right ) $. Furthermore, the probability density function (PDF) of a certified response is completely determined by its own velocity, regardless of other response quantities. Fundamentally, the PDEM generates a number of 2D $p_{X_{\bm a}}\left ( x,t \right )$ results from their deterministic 1D system responses $X \left ( \bm{a},t \right )$ using Eq. (13). These results are superimposed to approximate the desired $p_{X}\left ( x,t \right ) $ as a discrete realization of Eq. (14). The full PDEM algorithm is summarized in Algorithm~\ref{algo2}.\par
\begin{algorithm}
\caption{Computing $p_{X}\left ( x,t \right ) $ with PDEM }\label{algo2}
\begin{algorithmic}[1]
\STATE Pick out representative points $\bm{a}_q\left ( q=1,2,\cdots ,N_{sel} \right)$ in $\mathcal A$
\FOR{$q = 1 \to N_{sel}$}
	\STATE Carry out deterministic time integration on the 4-storey building, yielding the velocity $\dot{X} \left ( \bm{a}_q,t \right )$.
	\STATE Introducing the evaluated $\dot{X} \left ( \bm{a}_q,t \right )$, solve Eq. (13) with the finite difference method, yielding $p_{X_{\bm {a}_q}}\left ( x,t \right )$
\ENDFOR
\STATE $p_{X}\left ( x,t \right ) =\frac{1}{N_{sel}}\sum_{q=1}^{N_{sel}} p_{X_{\bm {a}_q}}\left ( x,t \right ) $ 
\end{algorithmic}
\end{algorithm}

$X$ in Eq. (13) can be any desired physical quantity from displacement, velocity, and acceleration to the main stress. However, it should be emphasized that solving the 1D convection PDE in Eq. (13) is time-consuming, although multiple quantities can be solved in parallel. Therefore, the PDEM can only provide the evolution results of PDF ($p_{X}\left ( x,t \right ) $) for a limited number of $X$. In chapter ~\ref{Case3Chapter}, we will focus on the main stress of concrete for a few locations on a large-scale structure (see Fig.~\ref{Case3_Probability_Field}b).\\

\section{Results}
\subsection{Application demonstration on a toy example}
\begin{figure*}
\centering
\includegraphics[width=12cm]{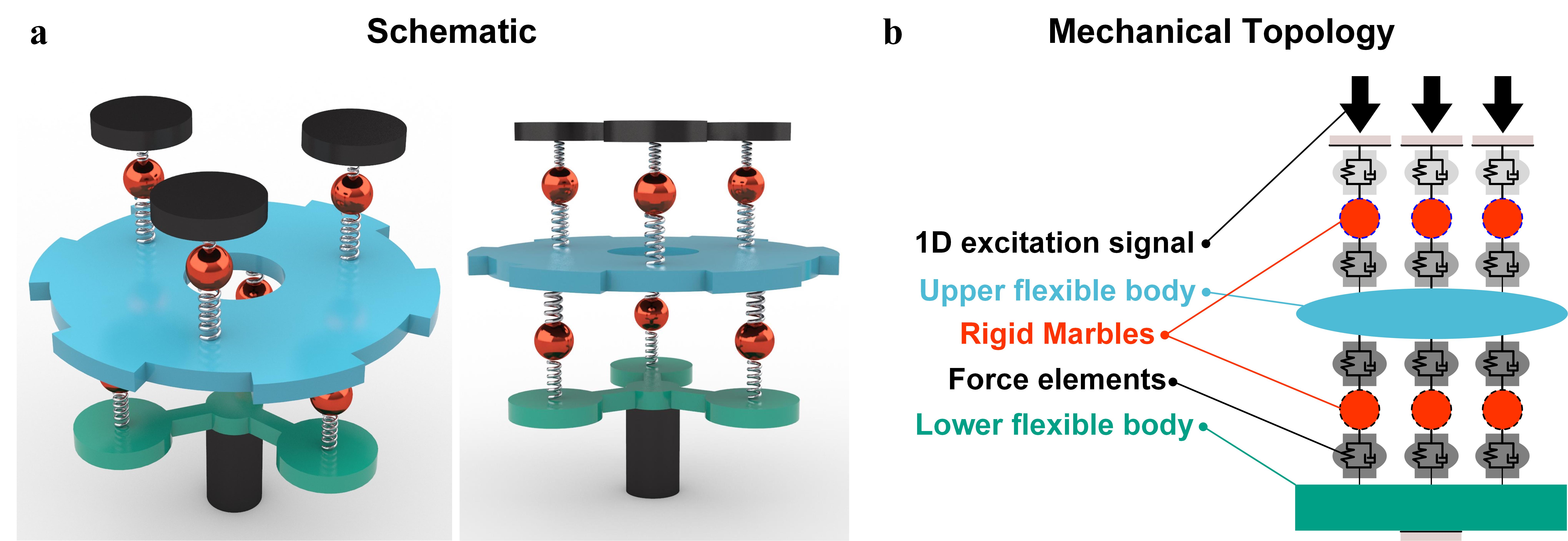}
\caption{\textbf{$\mid$ Schematic and topology of the toy example. a,} 3D schematic. \textbf{b,} Topology.}
\label{Case1_Schematic_and_Topology}
\end{figure*}

To demonstrate the flexibility of PINO-CDE, and to validate its capability to guarantee the accuracy for both solutions and derivatives, a typical mechanical system is used as a toy example. This example consists of two flexible rubber components with different elastic modules, and six rigid steel marbles, all of which are connected to each other via a spring-dashpot mechanism. In the mechanical module, 15 and 10 modes were considered for the upper and lower rubber components respectively. We trained the PINO-CDE to output solutions for all 43 DOFs, considering variable marble mass, spring-dashpot properties, and stochastic excitations. All data (solutions and derivatives) were first normalized to the dataset (with standardization) before their relative L2 losses ($rLSE$) were calculated to evaluate the performance of the PINO-CDE.\par

\begin{figure*}
\centering
\includegraphics[width=15cm]{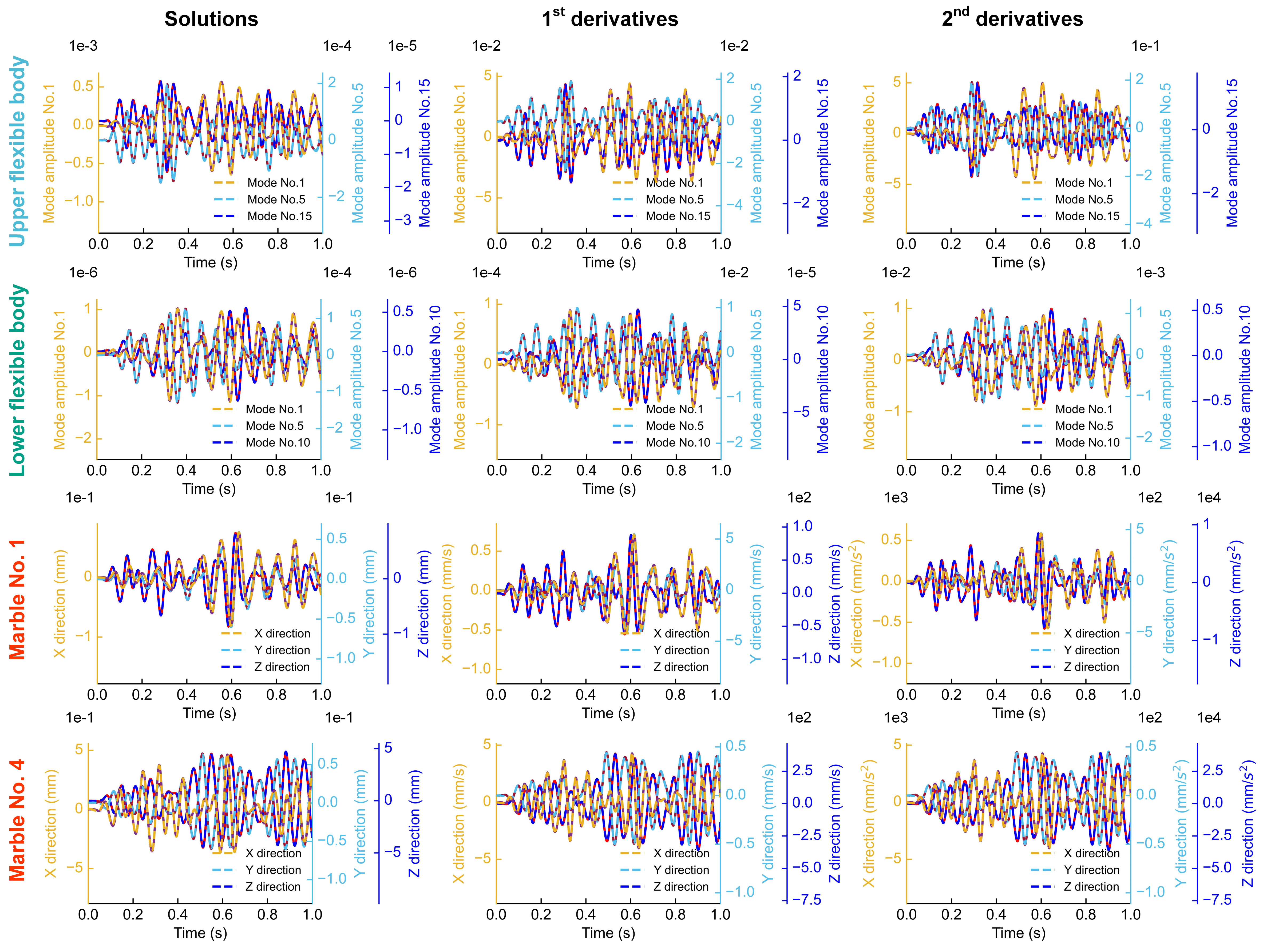}
\caption{\textbf{$\mid$ PINO-CDE performance for 1D signals.} The first to fourth rows of the composite figure correspond to the upper flexible body, lower flexible body, leftmost steel ball in the upper layer, and leftmost steel ball in the lower layer of the toy model, respectively. The first to third columns of the composite figure correspond to the solutions (displacement responses) of the coupled ordinary differential equation group as well as the first and second derivatives (velocity and acceleration responses) of the solutions. Each subfigure displays multiple DOFs for the represented objects, including multiple modes for the flexible components and vibrations in the X, Y and Z directions for the rigid objects. Additionally, in each subfigure, the solid lines present the ground truth solved through numerical integration, while the dashed lines display the ourput of PINO-CDE on a data pair from the test set.}
\label{Case1_Mode_Performance}
\end{figure*}

\begin{figure*}
\centering
\includegraphics[width=15cm]{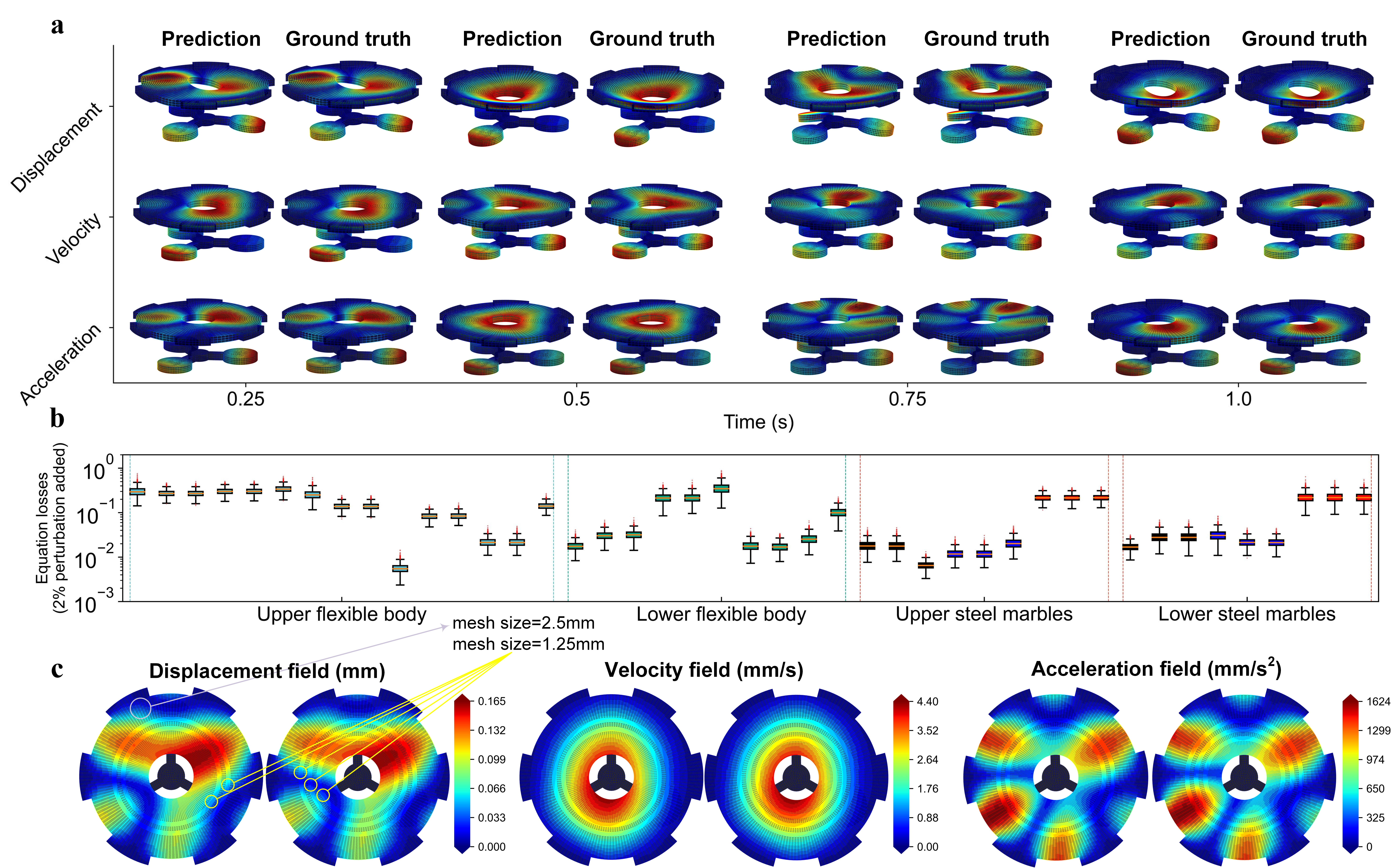}
\caption{\textbf{$\mid$ PINO-CDE performance for the recovered 3D responses. a,} Recovered 3D solution and derivative fields for the flexible components. The results for displacement fields are visualized with a scale factor of 50. \textbf{b,} Statistical results of equation losses for different ODEs with a 2$\%$ perturbation error added to their ground truth. \textbf{c,} Mesh-independence verification of PINO-CDE. Recovered solution and derivative fields from two different mesh grids are compared. The annular areas in both grids are meshed with a finer size.}
\label{Case1_Recovered_Performance}
\end{figure*}

After 300 epochs, the 1D responses of mode amplitudes for the two rubber components agreed well with the ground truth obtained using numerical integration; the same was observed for the vibration responses of the steel marbles (overall $rLSE=3.96\%$). In addition, PINO-CDE demonstrated good accuracy on the derivatives of the solutions, with the $rLSE$ of the $1^{st}$ and $2^{nd}$ derivatives being 3.73$\%$ and 4.47$\%$, even though the ground truth for derivatives was not provided during training. The displacement, velocity, and acceleration fields for the two flexible components were recovered at different positions along the timeline. As shown in Fig.~\ref{Case1_Recovered_Performance}a, the responses in the three-dimensional space demonstrated sufficient accuracy, suggesting that PINO-CDE is potentially useful in GUI-based simulation applications for engineering design. We also provide a video for the complete animation at our \href{https://www.youtube.com/@qunfangding1898/featured}{Youtube channel} .\par

Next, Fig.~\ref{Case1_Recovered_Performance}b shows the equation loss statistics for each individual equation when 2$\%$ perturbation error is added to the ground truth. This operation was to simulate the learning process, and the variation in the magnitude of different equation losses indicates that the gradients PINO-CDE harvested from different equations and data pairs were uneven during training. We discuss the effect of this gradient imbalance phenomenon on the training quality in Subsection ~\ref{Analysis on training quality}.\par 
Finally, the mesh-independence of PINO-CDE is verified in Fig.~\ref{Case1_Recovered_Performance}c. The annual areas in both grids were meshed to a finer size. The trained PINO-CDE can directly use new mode shape vectors to recover the responses, requiring no additional training, which differentiates PINO-CDE from other mesh-dependent deep learning methods \cite{bib29, bib30, bib31}. In practice, one can simply use element shape functions to obtain responses at desired positions, just like the finite element method \cite{bib32}.\\

\subsection{Practical application of VTCD}
\begin{figure*}
\centering
\includegraphics[width=15cm]{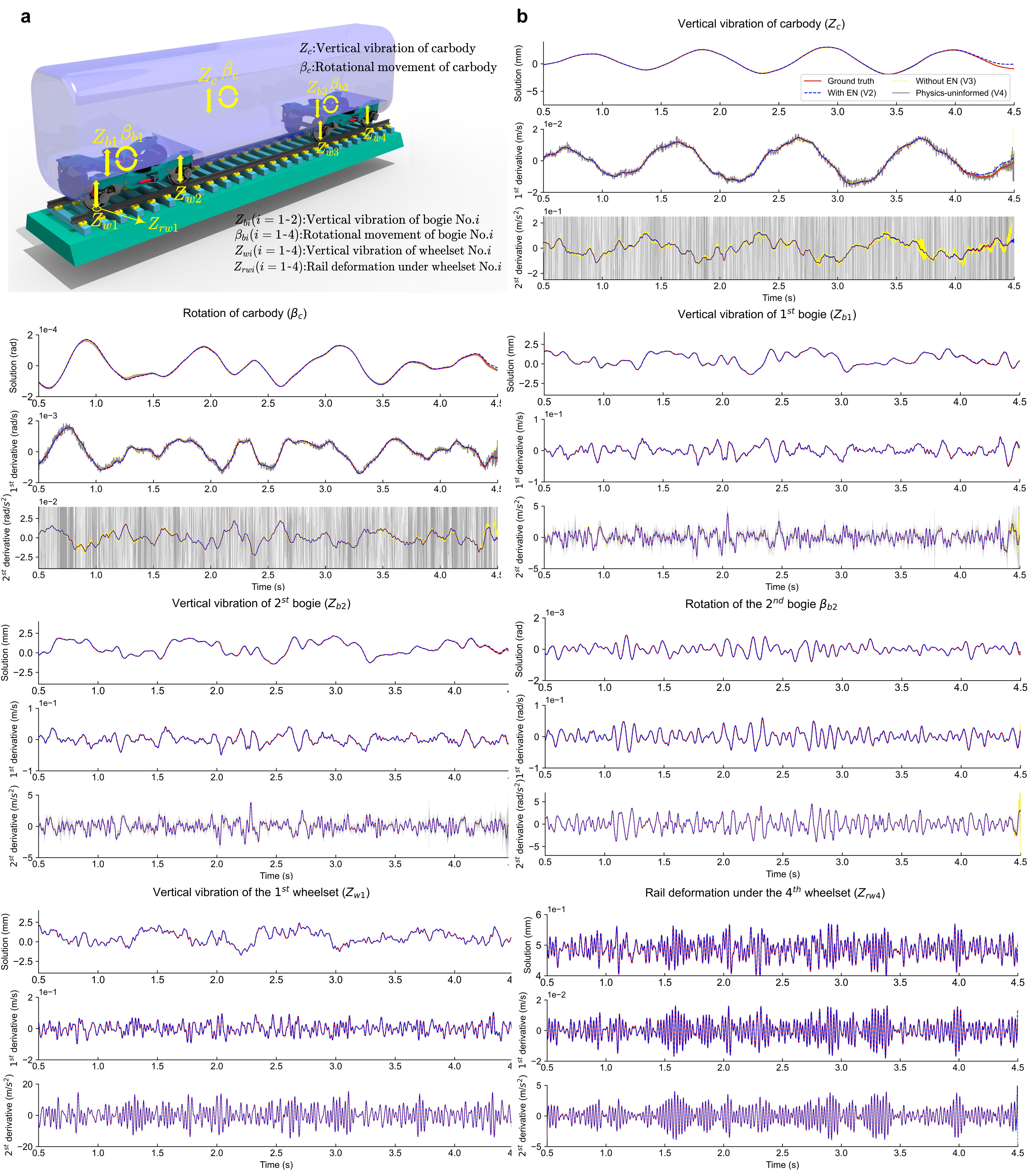}
\caption{\textbf{$\mid$ Application demonstration of VTCD. a,} Schematic for the train-track coupled system. \textbf{b,} Performance of PINO-CDE on the dynamic responses of the vehicle-track coupled system. The solutions and their derivatives are visualized for 7 out of 14 DOFs. Definitions for the three cases (V2, V3, and V4) demonstrated here are listed in Table~\ref{tab1}.}
\label{Case2_Signal_Performance}
\end{figure*}

As major transportation arteries in many countries, railways play an important role in social and economic development, which has stimulated the study of VTCD \cite{bib33, bib34}. In the past 20 years, the theory of VTCD has been verified through numerous site experiments \cite{bib35}. Currently, VTCD is widely used in many aspects of railway engineering, including integrated designs of modern rolling stocks and track structures, safety evaluation of new lines before operation, and engineering education. These practical demands have facilitated the development of corresponding software modules, such as Simpack$^\circledR$ and UM$^\circledR$, with which we intend to compete using the PINO-CDE.\par
\begin{figure*}
\centering
\includegraphics[width=15cm]{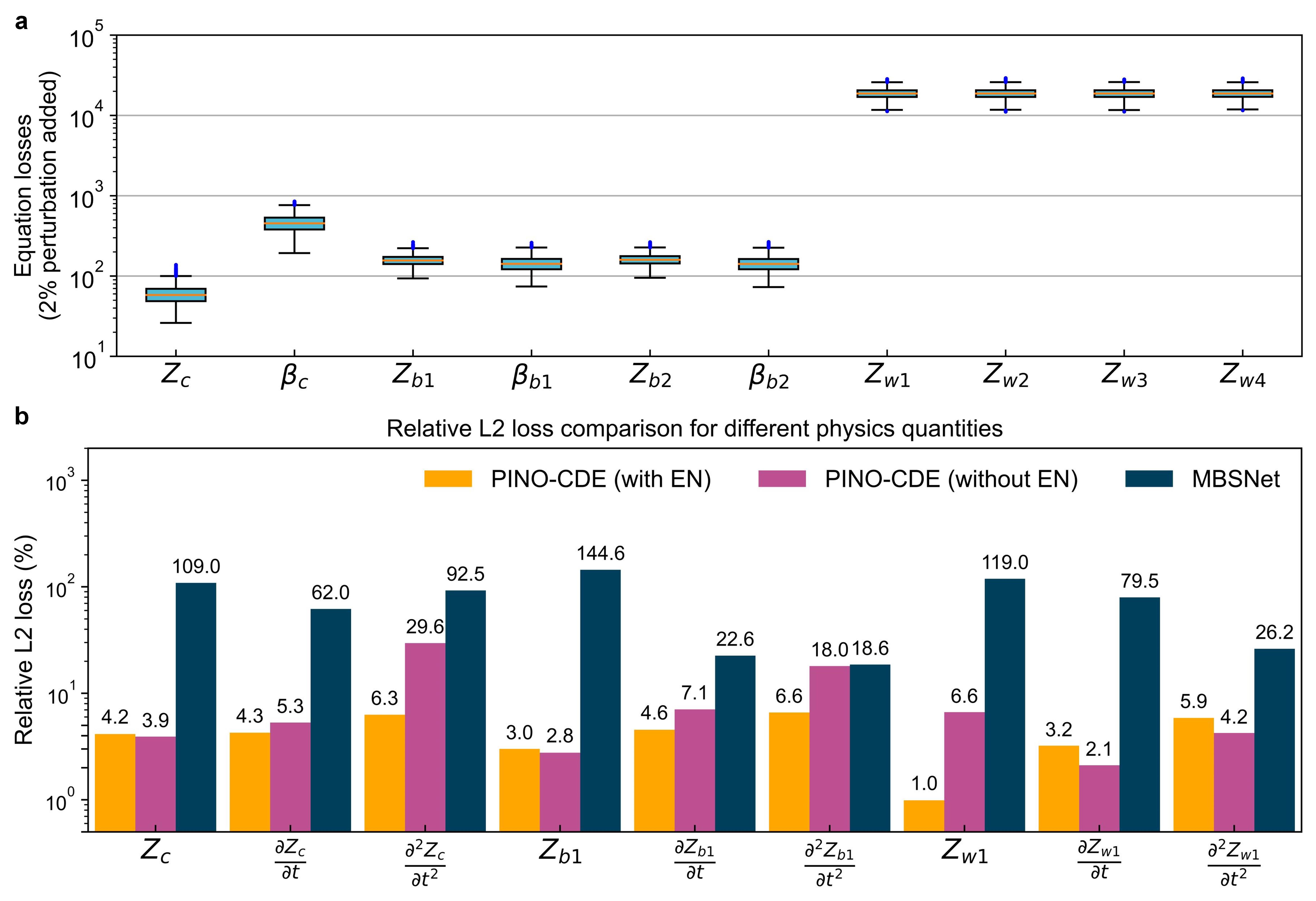}
\caption{\textbf{$\mid$ Analysis on PINO-CDE's performance on VTCD.} \textbf{a,} Statistical results of equation losses for different ODEs with a 2$\%$ perturbation error added to their ground truth. \textbf{b,} Performance comparison among PINO-CDE trained with EN (V2), PINO-CDE trained without EN (V3), and MBSNet. The performance data of MBSNet was extracted from \cite{bib14}.}
\label{Case2_Analysis_on_PINO-CDE_Performance_for_VTCD}
\end{figure*}

In this example, all properties of the running vehicle were considered to be variable, including suspension parameters, the distance between bogies, the distance between wheelsets, and running speed. For the track structure, the stiffness and damping of the rail pads, along with stochastic rail irregularities on the surface, were also variable. Once trained, PINO-CDE can instantly output dynamic responses for the vehicle-track system (10 DOFs for the vehicle system and 4 DOFs for rail deformations under wheelsets) considering arbitrary excitations and parameter configurations. As shown in Fig.~\ref{Case2_Analysis_on_PINO-CDE_Performance_for_VTCD}a, the output results (V2 in Table~\ref{tab1}) achieved good accuracy. Specifically, system displacement, velocity, and acceleration achieved 4.80$\%$, 4.23$\%$, and 4.30$\%$ $rLSE$ after 300 epochs.\par 
Compared with numerical integration, the main advantage of PINO-CDE lies in its efficiency. For a 5 s dynamic response of a train running at 350 km/h, UM$^\circledR$ and Simpack$^\circledR$ take 18 and 23 s, respectively. In contrast, performing a forward prediction of 10,000 cases with PINO-CDE takes less than 0.02 s, indicating a magnitude improvement of three orders in computational efficiency. Furthermore, solving the VTCD with PINO-CDE only requires sufficient memory on the computer, whereas numerical integration methods are notorious for requiring a considerable amount of CPU processing power.\par

On the other hand, PINO-CDE’s performance is more elegant in terms of both functionality and precision than the existing deep-learning based approach MBSNet \cite{bib14}. As shown in Fig.~\ref{Case2_Analysis_on_PINO-CDE_Performance_for_VTCD}d, PINO-CDE outperformed MBSNet by at least one order of magnitude in accuracy for most DOFs. In addition, PINO-CDE not only provides solutions but also guarantees the accuracy of their derivatives, which is much more challenging than just providing all results, as is done by MBSNet.\\

\subsection{Practical application of reliability analysis to a large-scale structure under seismic attack}
\label{Case3Chapter}
\begin{figure*}
\centering
\includegraphics[width=15cm]{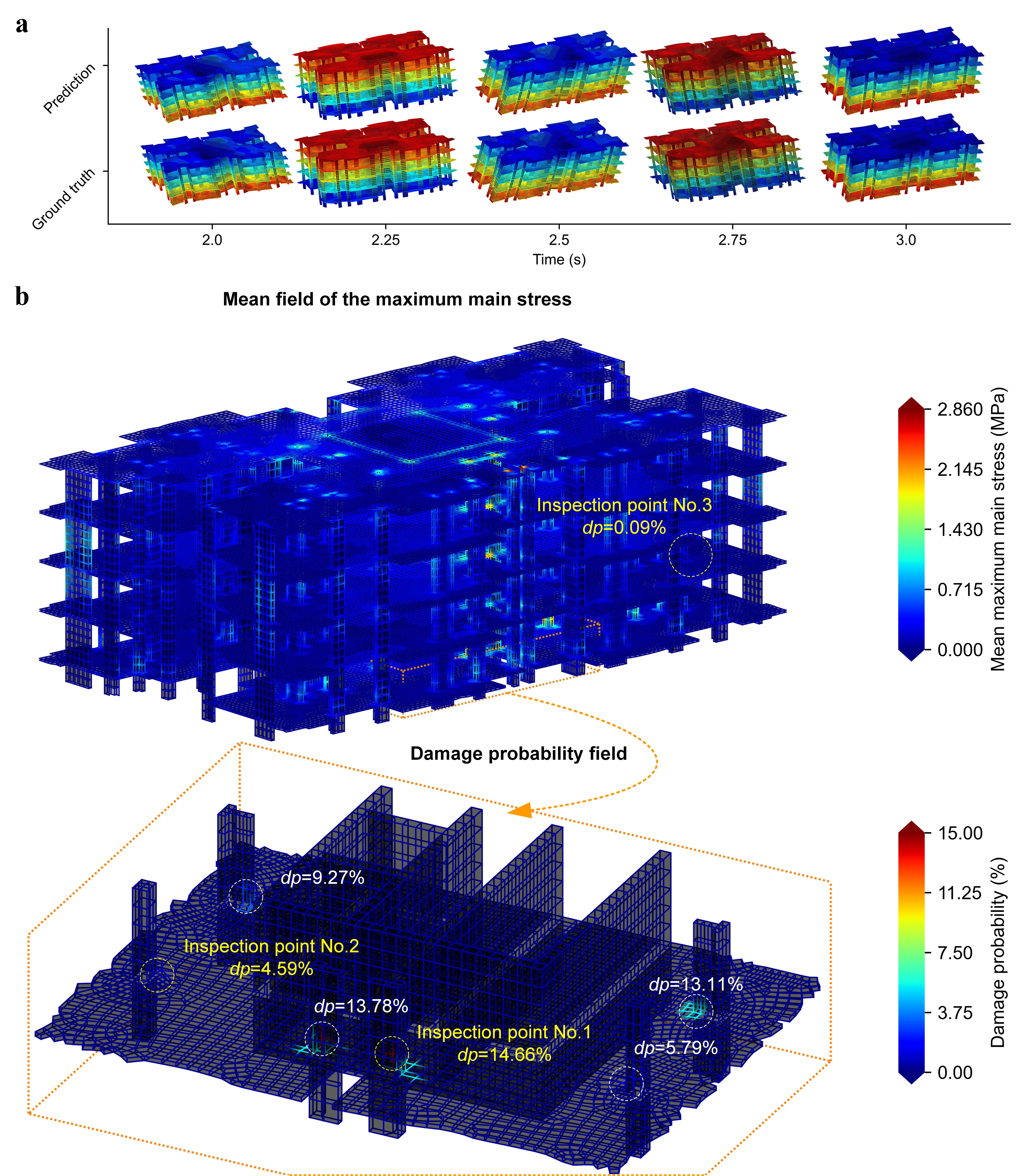}
\caption{\textbf{$\mid$ Application demonstration of reliability assessment (uncertainty propagation) on a large-scale structure. a,} Recovered displacement field of a four-storey building under seismic attack. \textbf{b,} The mean field of maximum main stress is shown on the left and the damage probability field is on the right.}
\label{Case3_Probability_Field}
\end{figure*}

To further demonstrate the potential of the PINO-CDE for probabilistic engineering, we performed a reliability assessment (uncertainty propagation) for a four-storey building modeled with 67881 solid elements (135426 DOFs), as shown in Fig.~\ref{Case3_Probability_Field}. The task of a probabilistic engineer is to evaluate the aseismic reliability by solving the probability characteristics of its dynamic response under a 3D stochastic seismic attack. The 3D seismic excitations applied to the foundation of the building were generated using the Kanai-Tajimi spectrum. The earthquake acceleration signal in each direction was generated using a spectrum method with one random phase \cite{bib48}, forming a 3D random parameter space $\mathcal A$ for this example.

This task can be achieved using the Monte Carlo method (MCM), that is, solving a large number of cases considering the stochastic nature of earthquakes, and then statistically obtaining the damage probability. However, this approach is practically infeasible owing to the power-hungry nature of MCM. For this particular building, it took approximately 1 h to solve a 5-second seismic response on an Intel Core i9-11900K CPU, implying that it would take more than a year to solve 50,000 cases with five computers.\par
Bypassing the computational consumption of the MCM is one of the main research goals in the field of probabilistic engineering. Since the study of Brownian motion by Einstein (1905), many approaches to this problem have been proposed, such as the Liouville equation, FPK equation, and Dostupov-Pugachev equation \cite{bib37}. We used the probability density evolution method (PDEM) \cite{bib36, bib37, bib38} (workflow listed in subsection \ref{Review_PDEM}) as the baseline for the PINO-CDE to compete with, considering its popularity and successful applications on many engineering structures, such as buildings \cite{bib39}, slopes \cite{bib40}, tie-back walls \cite{bib41}, and bridges \cite{bib42, bib43}. Fundamentally, the PDEM produces the probability density function (PDF) results of the desired response over time by superimposing the evolution results of a smaller number of cases (compared to MCM). We used a uniform design \cite{bib47} based on number theory to extract 499 representative points in $\mathcal A$ uniformly, which we believe is sufficient compared with similar research items. The evolution result of each case is obtained by solving a 1D convection equation that describes the flow of the deterministic responses in the probability space.\par
We trained the PINO-CDE to learn the mapping between seismic excitations and the structural response on a small dataset with 150 data pairs. After verifying the PINO-CDE on a test set with 30 data pairs, we let it perform fast forward prediction under 49,999 stochastic 3D seismic waves, and subsequently recovered the stress field to obtain the damage probability statistically. This process shares the same philosophy as MCM, except that the computational cost is significantly reduced with the help of the PINO-CDE. Fig.~\ref{Case3_Probability_Field}a visualizes the recovered displacement fields generated with the output mode amplitude responses, in which 200 modes were considered. The PINO-CDE prediction achieved good accuracy compared with the ground truth, with a $rLSE$ of 4.81$\%$ after 300 epochs.\par
\begin{figure*}
\centering
\includegraphics[width=15cm]{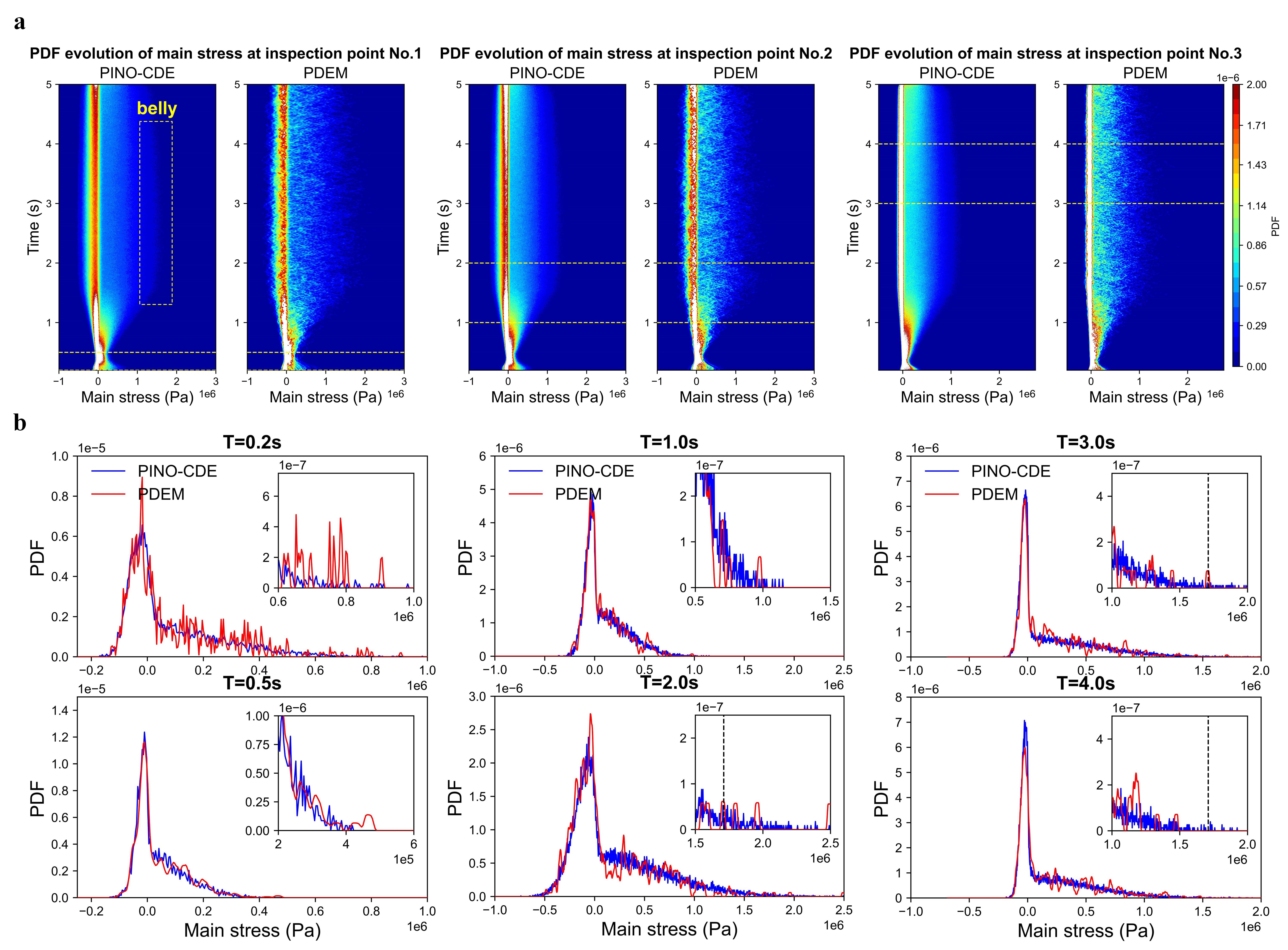}
\caption{\textbf{$\mid$ Comparison between PINO-CDE and PDEM. }\textbf{a,} Probability density evolution results for different inspection points. We refer to the right edge area with higher main stress as the belly area. \textbf{b,} Sliced observations of probability density evolution surfaces at different times. The observation positions on the time axis are visualized as yellow dash lines in sub-figure a. $dp^*$ denotes the damage probability at the specific time.}
\label{Case3_PDEM}
\end{figure*}

The probability results statistically obtained from 49,999 predictions using the trained PINO-CDE are shown in Fig.~\ref{Case3_Probability_Field}b and Fig.~\ref{Case3_PDEM}a-b. Through these results, we observe that PINO-CDE outperforms PDEM in the following aspects.\\
\textbf{1). PINO-CDE allows engineers to observe the entire probability field of the structure.}\par
Looking at Fig.~\ref{Case3_Probability_Field}b, the hotter areas on the left have larger mean values of tensile stress over a total of 49,999 data pairs. On the right, $dp$ denotes the probability of concrete cracking and fracturing, that is, the probability that the main stress exceeds the tensile strength (1.71 MPa) during an earthquake. Using this figure, engineers can easily determine the vulnerable areas of the structure and the damage probability at different locations, providing decision-makers with a basis for evaluating and optimizing designs. In contrast, it is impossible to solve the probability evolution results for all DOFs using the PDEM because solving the 1D convection equation is time-consuming. Therefore, PDEM can only provide results for a limited number of observation positions; that is, Fig.~\ref{Case3_Probability_Field}b cannot be generated with the PDEM. Similarly, classic surrogate models typically only target on individual mechanical indicators \cite{bib53}. Unless combined with data dimensionality reduction techniques, they cannot be readily used to reconstruct the complete probability field of the structure \cite{bib54}.\\
\textbf{2). PINO-CDE provides probability evolution results with a higher resolution.}\par
As shown in Fig.~\ref{Case3_PDEM}a, the probability evolution results produced by the PDEM are rough and uneven compared with those of PINO-CDE. A closer examination of Fig.~\ref{Case3_PDEM}b reveals many isolated small probability hills in the belly area from the results generated with the PDEM. This phenomenon has been reported in many papers \cite{bib40, bib44, bib45}, but to the best of our knowledge, it has not been discussed. The high-resolution results of PINO-CDE indicate that the roughness of the evolution results and the isolated hills in the belly area are not inherent properties of the probabilistic results per se, but rather a manifestation of the PDEM’s insufficient resolution. This advantage of PINO-CDE has practical implications for reliability assessment. Considering Fig.~\ref{Case3_PDEM}b as an example, at T=4.0 s, $dp^*$ calculated by the PINO-CDE is $0.03\%$ whereas that calculated by the PDEM is less than $0.01\%$, suggesting that the PDEM can be unconservative.\\
\textbf{3). The PINO-CDE was three times more efficient.}\par
PINO-CDE took only 37 h on five threads, including generating training data (30 h), training (3 h), and post-processing (4 h). Contrastingly, PDEM took 149 h, including computing 499 cases (100 h), evolving the results (48 h), and post-processing (1 h). PINO-CDE demonstrated a 302.7$\%$ increase in computational efficiency.\\

\subsection{Analysis on training quality}\label{Analysis on training quality}
\begin{figure*}
\centering
\includegraphics[width=14cm]{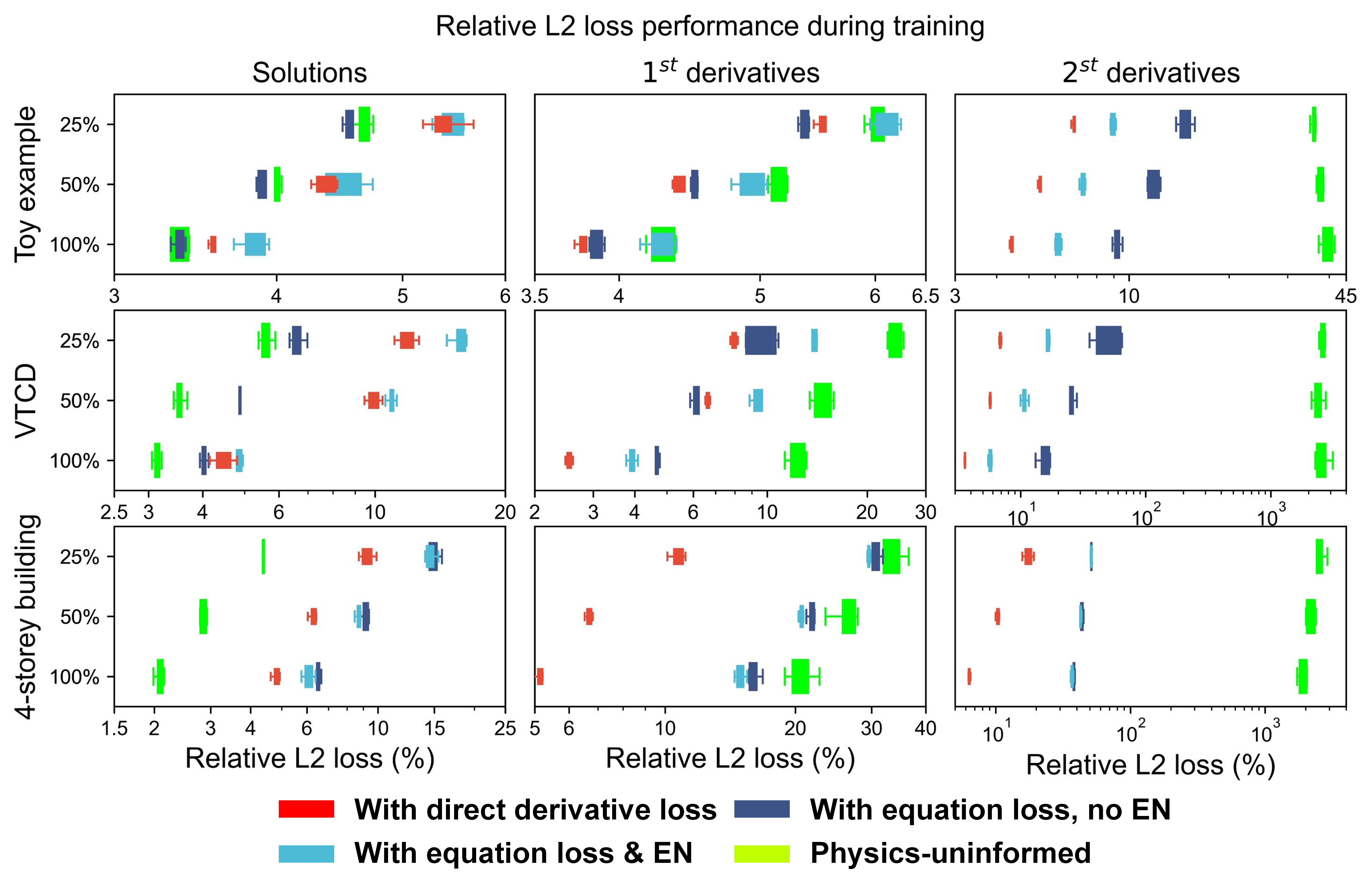}
\caption{\textbf{$\mid$ $rLSE$ performances for different loss function combinations. } $rLSE$ for CDEs solutions and their derivatives are visualized at 25$\%$, 50$\%$, and 100$\%$ of the training process. The first to third rows correspond to the toy example, VTCD, and the reliability assessment on the four-storey building. Each color for the boxes represents an individual case, corresponding to Table~\ref{tab1}.}
\label{Training_Visualization}
\end{figure*}

\begin{table*}[h]
\begin{center}
\begin{minipage}{\textwidth}
\caption{Performance analysis under different loss function combinations. (average $rLSE$ over multiple training).}\label{tab1}
\begin{tabular*}{\textwidth}{@{\extracolsep{\fill}}lcccccccccc@{\extracolsep{\fill}}}
\toprule%
Index & \multicolumn{2}{@{}c@{}}{Network} & \multicolumn{3}{@{}c@{}}{Loss function} & \multicolumn{4}{@{}c@{}}{$rLSE$ $(\%)$} \\\cmidrule{2-3}\cmidrule{4-6}\cmidrule{7-10}%
& PI & EN & $\mathcal L_{eq}$ & $\mathcal L_{dde}$ & $\mathcal L_{veq}$ & Solutions & $1^{st}\;De$s & $2^{st}\;De$s & Average \\
\midrule
\multicolumn{10}{@{}c@{}}{Toy example}\\
\midrule
\cellcolor[rgb]{0.302, 0.7333, 0.8353}T1 & $\checkmark$ & $\checkmark$ & * &   &   & 3.85 & 4.28 & 6.14 & 4.76\\
T2 & $\checkmark$ & $\checkmark$ & * & * &   & 3.54 & 3.73 & 4.47 & 3.91\\
T3 & $\checkmark$ & $\checkmark$ & * &   & * & 3.96 & 4.39 & 6.03 & 4.79\\
\cellcolor[rgb]{0.902,0.2941,0.2078}T4 & $\checkmark$ & $\checkmark$ & * & * & * & 3.60 & 3.80 & 4.46 & 3.95\\
\cellcolor[rgb]{0.2353, 0.3294, 0.5333}\textcolor{white}{T5} & $\checkmark$ &   & * &   &   & 3.34 & 3.86 & 9.24 & 5.49\\
T6 & $\checkmark$ &   & * &   & * & 3.37 & 3.79 & 7.81 & 4.99\\
\cellcolor{lime}T7 &   &   &   &   &   & 3.37 & 4.28 & 39.51 & 15.72\\
\midrule
\multicolumn{10}{@{}c@{}}{Vehicle-track coupled dynamics}\\
\midrule
\cellcolor[rgb]{0.902,0.2941,0.2078}V1 & $\checkmark$ & $\checkmark$ & * & * &   & 4.22 & 2.42 & 3.42 & 3.35\\
\cellcolor[rgb]{0.302, 0.7333, 0.8353}V2 & $\checkmark$ & $\checkmark$ & * & + &   & 4.80 & 3.56 & 5.25 & 4.54\\
\cellcolor[rgb]{0.2353, 0.3294, 0.5333}\textcolor{white}{V3} & $\checkmark$ &   & * & + &   & 4.12 & 4.77 & 15.14 & 8.00\\
\cellcolor{lime}V4 &   &   &   &   &   & 3.14 & 12.32 & 2584.13 & 866.53\\
\midrule
\multicolumn{10}{@{}c@{}}{Reliability assessment for a 4-storey building}\\
\midrule
\cellcolor[rgb]{0.902,0.2941,0.2078}R1 & $\checkmark$ & $\checkmark$ & * & * & * & 4.81 & 5.13 & 6.37 & 5.44\\
\cellcolor[rgb]{0.302, 0.7333, 0.8353}R2 & $\checkmark$ & $\checkmark$ & * & + & * & 6.03 & 14.91 & 36.90 & 19.28\\
\cellcolor[rgb]{0.2353, 0.3294, 0.5333}\textcolor{white}{R3} & $\checkmark$ &   & * & + &   & 6.74 & 16.04 & 38.04 & 20.27\\
\cellcolor{lime}R4 &   &   &   &   &   & 2.62 & 23.51 & 2116.41 & 714.18\\
\midrule
\label{Case analysis}
\end{tabular*}
\footnotetext{In this table, PI marked with $\checkmark$ indicates the trained model is physics-informed, while EN marked with $\checkmark$ means equation normalization technique was used. For each experiment, loss components marked with * were used during training. For $\mathcal L_{dde}$ marked with +, we only provided the ground truth of derivatives for a small period of time (0.025s) on the edges of the time domain as a boundary constraint. $De$s stands for derivatives, and experiments with colored backgrounds are visualized in Fig.~\ref{Training_Visualization}. All data were normalized on corresponding datasets before their $rLSE$ were computed.}
\end{minipage}
\end{center}
\end{table*}

To further understand the key elements affecting training quality, we conducted a series of experiments, considering different network properties and loss function compositions, as shown in Table~\ref{tab1} and Fig.~\ref{Training_Visualization}. First, we note the importance of embedding physical knowledge for engineering dynamics. A naked neural operator as the backbone generally demonstrates slightly better performance on solutions but deteriorates dramatically on derivatives (T1 vs T6, V2 vs V4, R2 vs R4, Fig.~\ref{Case2_Signal_Performance}b). These results echo the finding in \cite{bib18}, emphasizing the necessity of embedding physics because the engineering community often values derivatives more than solutions.\par
Next, the benefits of using EN were noted by comparing T1 with T5, V1 with V3, and R2 with R3. EN enhances the performance of the PINO-CDE on derivatives at the cost of a slight deterioration of solutions. For VTCD, EN offered 1.21$\%$ and 9.89$\%$ improvement on $1^{st}$ and $2^{nd}$ derivatives, but only a 3.1$\%$ improvement on $2^{nd}$ derivatives was observed for the toy example. We reason that this is because the imbalance situation for VTCD is much worse, as observed from Fig.~\ref{Case1_Recovered_Performance}b and Fig.~\ref{Case2_Analysis_on_PINO-CDE_Performance_for_VTCD}a. Without EN, the gradients harvested from some ODEs (those with larger equation losses when the perturbation is added) are much larger than others, prohibiting PINO-CDE from learning mappings for all DOFs uniformly. For applications with higher precision requirements, direct $\mathcal L_{dde}$ is still necessary (which is generally wasted in existing studies) at the cost of more GPU occupation and generating ground truth for derivatives.\par 
Finally, the advantages of using virtual loss can be quantified by comparing T1 to T3, and T2 to T4. The benefits of using virtual loss $\mathcal L_{veq}$ are limited, possibly because their training datasets are already large enough. This naturally raises a question of whether PINO-CDE can be trained without providing any ground truth data (only with $\mathcal L_{eq}$ and $\mathcal L_{veq}$). We verify this idea by providing an additional example in section A.5.\\

\section{Hyperparameter analysis and fine-tuning}\label{Hyperparameter analysis and fine-tuning}
The hyperparameters of PINO-CDE include the batch size, learning rate, as well as the width and depth of the Fourier convolution layers and fully-connected layers. In the previous sections, these hyperparameters were manually selected, necessitating a hyperparameter analysis to be conducted. Additionally, model fine-tuning should be performed to enable readers to accurately reproduce and apply PINO-CDE in the aforementioned scenarios. Hence, this section focuses on conducting a hyperparameter analysis to evaluate the sensitivity of PINO-CDE to different hyperparameters. Bayesian optimization will then be utilized to determine the optimal combinations of hyperparameters for the three cases mentioned above.\par
The results presented in Table ~\ref{Case analysis} clearly demonstrate the importance of considering equation losses to ensure satisfactory performance, particularly for the output derivatives. Consequently, our analysis focuses solely on two of the cases mentioned in Table ~\ref{Case analysis} for each scenario. The first case employs $\mathcal L_{dde}$ (T2, V1, and R1), while the second case only employs $\mathcal L_{eq}$ (T1, V2, and R2). During the initial stages of our research, we discovered that the model's performance showed minimal sensitivity to the batch size. Therefore, we set the batch size to the maximum value allowed by GPU occupancy restrictions. We utilized the Adam algorithm as the optimizer and manually set the learning rate decay schedule for each case. Hence, our analysis will focus on investigating the width and depth of the Fourier convolution layers and fully-connected layers as the main targets.\par

\begin{figure*}
\centering
\includegraphics[width=14cm]{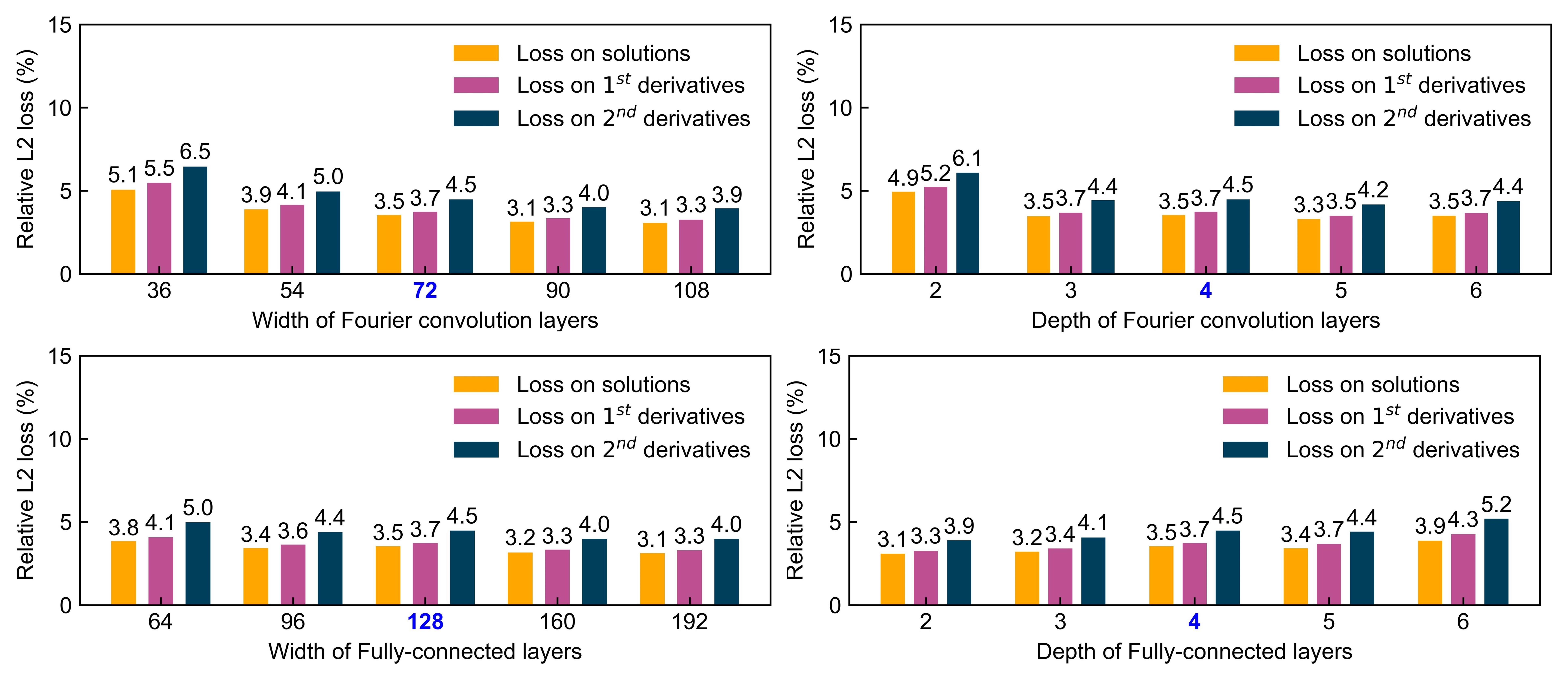}
\caption{\textbf{$\mid$ Hyperparameter analysis on the toy example (trained with $\mathcal L_{dde}$)}}
\label{HPA_toy_example}
\end{figure*}

\begin{figure*}
\centering
\includegraphics[width=14cm]{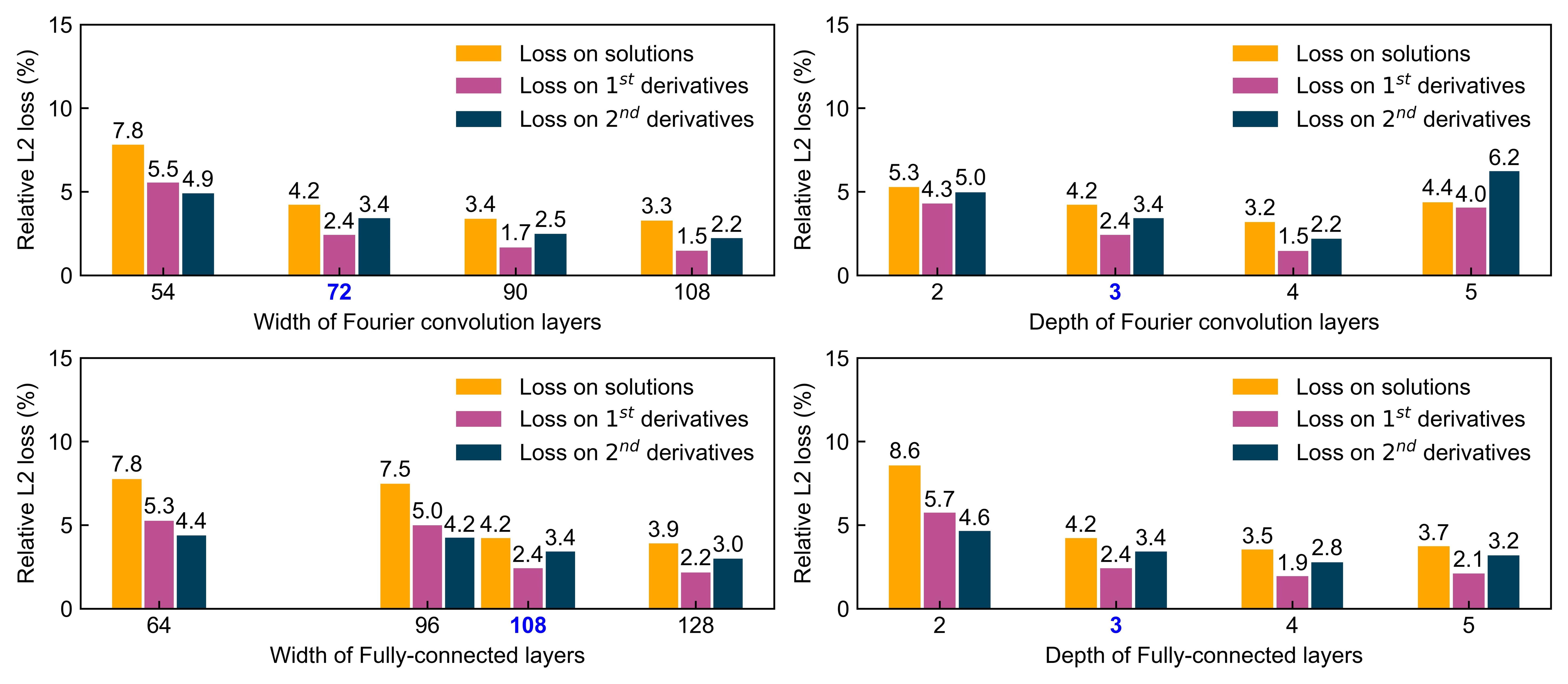}
\caption{\textbf{$\mid$ Hyperparameter analysis on vehicle-track coupled dynamics (trained with $\mathcal L_{dde}$)}}
\label{HPA_VTCD}
\end{figure*}

\begin{figure*}
\centering
\includegraphics[width=14cm]{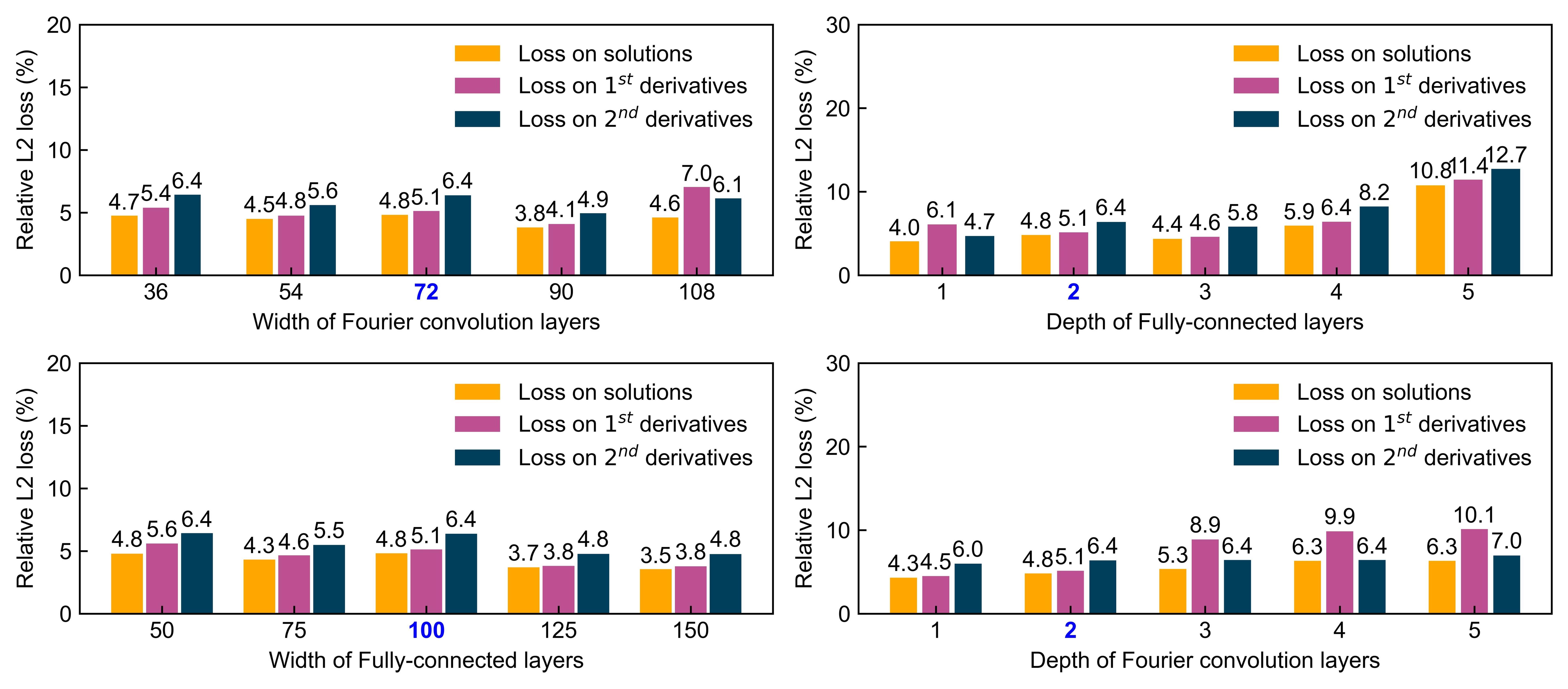}
\caption{\textbf{$\mid$ Hyperparameter analysis on the reliability assessment case (trained with $\mathcal L_{dde}$)}}
\label{HPA_BSA}
\end{figure*}

Figures Fig.~\ref{HPA_toy_example} to Fig.~\ref{HPA_BSA} present the hyperparameter analysis results for three cases. Bule labels on the x-axis denote the manual-determined hyperparameter values used in the above sections. These figures only show the results when using $\mathcal L_{dde}$ and $\mathcal L_{eq}$ simultaneously. When using only $\mathcal L_{eq}$, the performance trends of PINO-CDE is similar to the ones shown in the figures, but with relatively lower accuracy. The corresponding results can be found in the supplementary materials of this paper. From these figures, it can be observed that PINO-CDE exhibits limited sensitivity to the hyperparameters. Additionally, from these figures, it can be seen that increasing the width of Fourier convolution layers (Graph kernel network) has the most significant positive impact on accuracy improvement.To facilitate readers in reproducing the performance of PINO-CDE on the three cases mentioned above, we will now consider the aforementioned two types of equation loss configurations for Bayesian optimization. In this process, the toy case and the reliability evaluation case will undergo 15 and 10 optimizations, respectively. However, due to computational costs, the VTCD case will only undergo 8 optimizations. Furthermore, we will utilize the results of the aforementioned hyperparameter analysis as the initial known conditions for Bayesian optimization, effectively leveraging this information. The specific parameter boundaries, optimal parameter values, and corresponding optimal performance of PINO-CDE after Bayesian optimization are presented in Table ~\ref{Hyperparameter settings}.

\begin{table*}[h]
\begin{center}
\begin{minipage}{\textwidth}
\caption{Hyperparameter settings and optimial performances}\label{tab2}
\begin{tabular*}{\textwidth}{@{\extracolsep{\fill}}lcccccccc@{\extracolsep{\fill}}}
\toprule
\multicolumn{8}{@{}c@{}}{Hyperparameter settings} \\
\toprule
\multirow{2}{*}{Type} & \multirow{2}{*}{Hyperparameter} & \multicolumn{2}{@{}c@{}}{Toy example} & \multicolumn{2}{@{}c@{}}{VTCD} & \multicolumn{2}{@{}c@{}}{Reliability assessment} \\\cmidrule{3-8}
& & \multicolumn{2}{@{}c@{}}{Range} & \multicolumn{2}{@{}c@{}}{Range} & \multicolumn{2}{@{}c@{}}{Range} \\
\midrule
\multirow{4}{*}{Automatic} & Width of GKN & \multicolumn{2}{@{}c@{}}{{[}36, 108]}           & \multicolumn{2}{@{}c@{}}{{[}54, 108]}               & \multicolumn{2}{@{}c@{}}{{[}36, 144]}               \\
                           & Depth of GKN & \multicolumn{2}{@{}c@{}}{{[}2, 6]}             & \multicolumn{2}{@{}c@{}}{{[}2, 6]}             & \multicolumn{2}{@{}c@{}}{{[}1, 5]} \\
                           & Width of FCN & \multicolumn{2}{@{}c@{}}{{[}64, 192]}            & \multicolumn{2}{@{}c@{}}{{[}64, 128]}               & \multicolumn{2}{@{}c@{}}{{[}50, 144]}             \\
                           & Depth of FCN & \multicolumn{2}{@{}c@{}}{{[}2, 6]}             & \multicolumn{2}{@{}c@{}}{{[}2, 5]}               & \multicolumn{2}{@{}c@{}}{{[}1, 5]} \\
\midrule
\multirow{4}{*}{Manual} & Batch size & \multicolumn{2}{@{}c@{}}{100} & \multicolumn{2}{@{}c@{}}{30} & \multicolumn{2}{@{}c@{}}{15} \\
                & Epochs & \multicolumn{2}{@{}c@{}}{300} & \multicolumn{2}{@{}c@{}}{300} & \multicolumn{2}{@{}c@{}}{500} \\
                & Learning rate & \multicolumn{2}{@{}c@{}}{0.001} & \multicolumn{2}{@{}c@{}}{0.001} & \multicolumn{2}{@{}c@{}}{0.0005} \\
                & Decay steps & \multicolumn{2}{@{}c@{}}{75} & \multicolumn{2}{@{}c@{}}{50} & \multicolumn{2}{@{}c@{}}{100} \\
                & Decay ratio & \multicolumn{2}{@{}c@{}}{0.5} & \multicolumn{2}{@{}c@{}}{0.75} & \multicolumn{2}{@{}c@{}}{0.75}\\
\toprule
\multicolumn{8}{@{}c@{}}{Optimal performance} \\
\toprule
\multicolumn{2}{@{}c@{}}{ } & with $\mathcal L_{dde}$ & only $\mathcal L_{eq}$ & with $\mathcal L_{dde}$ & only $\mathcal L_{eq}$ & with $\mathcal L_{dde}$ & only $\mathcal L_{eq}$ \\
\midrule
\multicolumn{2}{@{}c@{}}{$rLSE$ on solutions} & 2.79 & 3.27 & 3.27 & 4.19 & 3.54 & 3.29 \\
\multicolumn{2}{@{}c@{}}{$rLSE$ on $1^{st}$ derivatives} & 2.96 & 3.63 & 1.46 & 3.61 & 3.78 & 7.08 \\
\multicolumn{2}{@{}c@{}}{$rLSE$ on $2^{nd}$ derivatives} & 3.55 & 5.59 & 2.22 & 5.74 & 4.75 & 23.88 \\
\midrule
\multicolumn{2}{@{}c@{}}{Width of GKN} & 108 & 104 & 108 & 96 & 72 & 108 \\
\multicolumn{2}{@{}c@{}}{Depth of GKN} & 3 & 3 & 3 & 4 & 2 & 2 \\
\multicolumn{2}{@{}c@{}}{Width of FCN} & 128 & 129 & 108 & 97 & 150 & 99 \\
\multicolumn{2}{@{}c@{}}{Depth of FCN} & 2 & 3 & 3 & 2 & 2 & 2 \\
\toprule
\label{Hyperparameter settings}
\end{tabular*}
\footnotetext{In this table, the hyperparameter bounds during the Bayesian optimization process are given for the three cases, as well as the optmal values and the corresponding performances of PINO-CDE. In the process of hyperparameter optimization, each case considered two equation loss configurations: one that simultaneously considered $\mathcal L_{dde}$ and $\mathcal L_{eq}$ (T2, V1, and R1), and another that only considered $\mathcal L_{eq}$ (T1, V2, and R2). Here, GKN denotes the Fourier convoluation layers (Graph Kernel Network) and FCN represents the Fully-connected layers.}
\end{minipage}
\end{center}
\end{table*}

\section{Discussion}
This study provides a proof of concept that physics-informed deep learning can successfully and efficiently regress solution operators for CDEs. Owing to the ever-increasing computing power in the last decade, engineering corporations and design institutes have accumulated huge simulation data but have conducted limited mining. By taking light from both mechanics and deep learning technologies, PINO-CDE allows engineers to model complex, practical systems with complicated boundary conditions and numerous details. More importantly, this can be achieved with a single network, instead of dozens or even hundreds as in the existing literature. This study, in turn, leads to a new concept that design institutes can train this framework for their specialized engineering objects to handle projects with different requirements, rather than modeling each project individually. In the future, we anticipate the development of software with GUI interfaces that will compete with existing simulation software.\par
To demonstrate its flexibility and feasibility, we provided a general toy example and two practical applications. For VTCD, PINO-CDE outperformed the commercial software and existing deep learning approaches in terms of efficiency and accuracy, respectively. We believe the better performance of the PINO-CDE compared to time series prediction methods such as CNN-LSTM (used in MBSNet) should be attributed to the stability of mechanical system responses in the frequency domain. In fact, finding the mappings between input and output functions in the frequency domain coincides with the idea of transfer function approaches \cite{bib2}. However, it is notoriously difficult to obtain transfer functions for complex structures using pure mechanical derivations, and our research has proven that deep learning can do better in this regard. Moreover, although the main target of our study is solid engineering dynamics, this framework can also be applied to other operator regression problems dealing with coupled differential equation groups. Different from existing approaches such as DeepM$\&$Mnet \cite{bib22, bib23}, we do not train multiple neural networks in advance. Instead, a single neural operator is used to output responses for all coupled physical quantities simultaneously. We believe this idea may provide a more concise approach to other disciplines such as multiphysics, which we will investigate in the future.\par
Furthermore, we demonstrated the potential of this framework for reliability assessments. We used this framework as a magnifying glass to observe the probabilistic characteristics of the dynamic structural responses. Thanks to the higher resolution brought by PINO-CDE, we found that the roughness and isolated probability hills are, in fact, caused by the insufficient resolution of the results solved from the probabilistic equation (PDEM), which has been observed in many studies. This may provide a new perspective for research in other disciplines since probabilistic and stochastic equations are widely used in physics \cite{bib49, bib50}, biology \cite{bib51}, and finance \cite{bib52}. In conclusion, this study integrates mechanics and deep learning-based operator regression technologies and enables engineers to solve complex systems efficiently and accurately. Although only a few examples are demonstrated here, we expect PINO-CDE to be broadly applicable to other structural and probabilistic engineering tasks.\\ 

\section*{CRediT authorship contribution statement}
\textbf{Wenhao Ding:} Conceptualization, Methodology, Software, Validation, Writing – original draft. \textbf{Qing He:} Formal analysis, Writing - review $\&$ editing. \textbf{Hanghang Tong:} Formal analysis - review $\&$ editing, Supervision. \textbf{Qingjing Wang:} Formal analysis, Writing - review $\&$ editing. \textbf{Ping Wang:} Supervision, Funding acquisition - review $\&$ editing.

\section*{Declaration of Competing Interest}
The authors declare that they have no known competing financial interests or personal relationships that could have appeared to
influence the work reported in this paper.

\section*{Declaration of generative AI and AI-assisted technologies in the writing process}
During the preparation of this work the authors used ChatGPT in order to improve readability in section ~\ref{Hyperparameter analysis and fine-tuning}. After using this tool, the authors reviewed and edited the content as needed and take full responsibility for the content of the publication.

\section*{Acknowledgements}
This work was funded by the National Natural Science Foundation of China (NSFC) under Grant Nos. U1934214 and 51878576.

\appendix
\section {Demonstration on training without data}
\begin{figure}
\centering
\includegraphics[width=.4\linewidth]{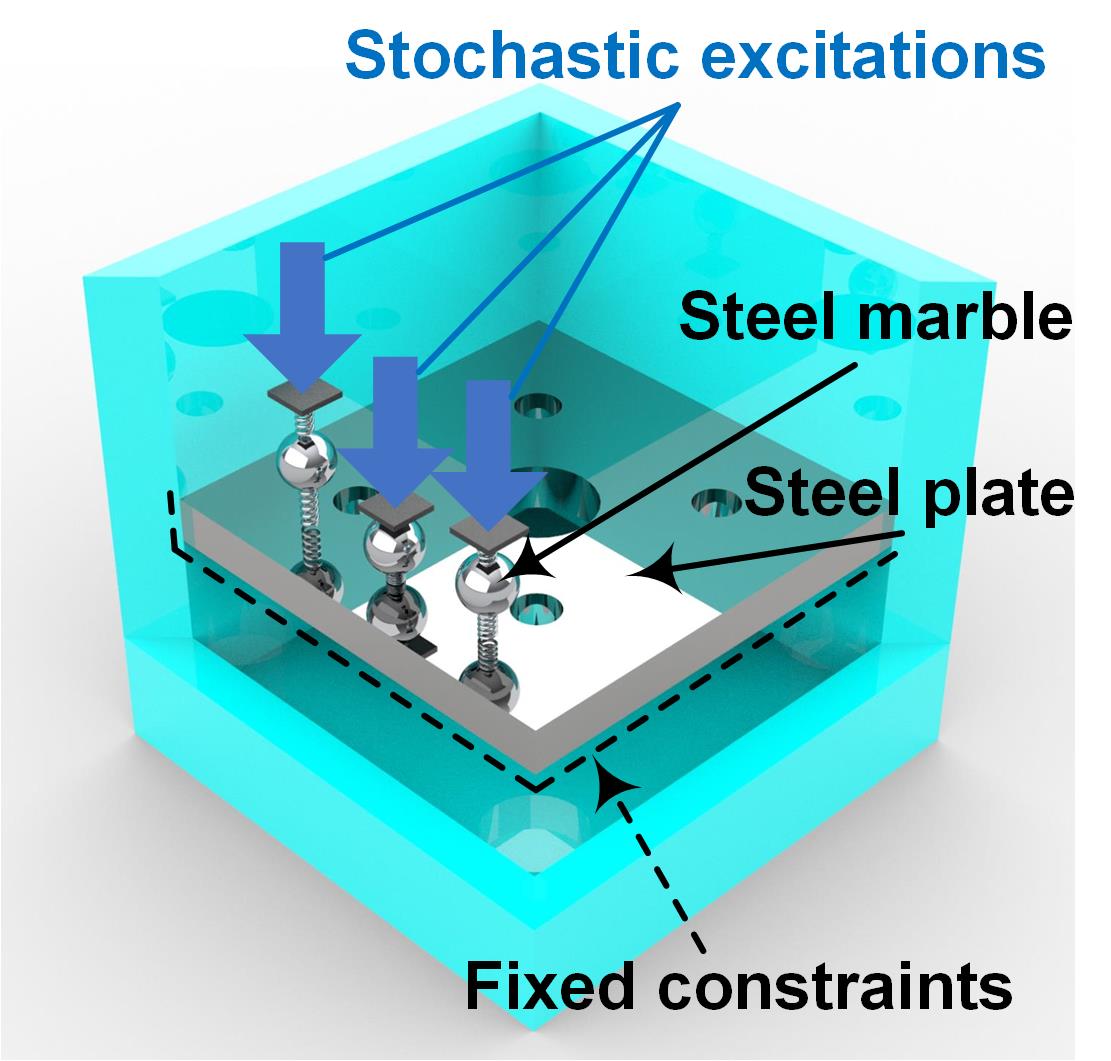}
\caption{\textbf{Schematic of the simple mechanical structure.}}
\label{Non-data_Schematic}
\end{figure}

\begin{figure*}
\centering
\includegraphics[width=.9\linewidth]{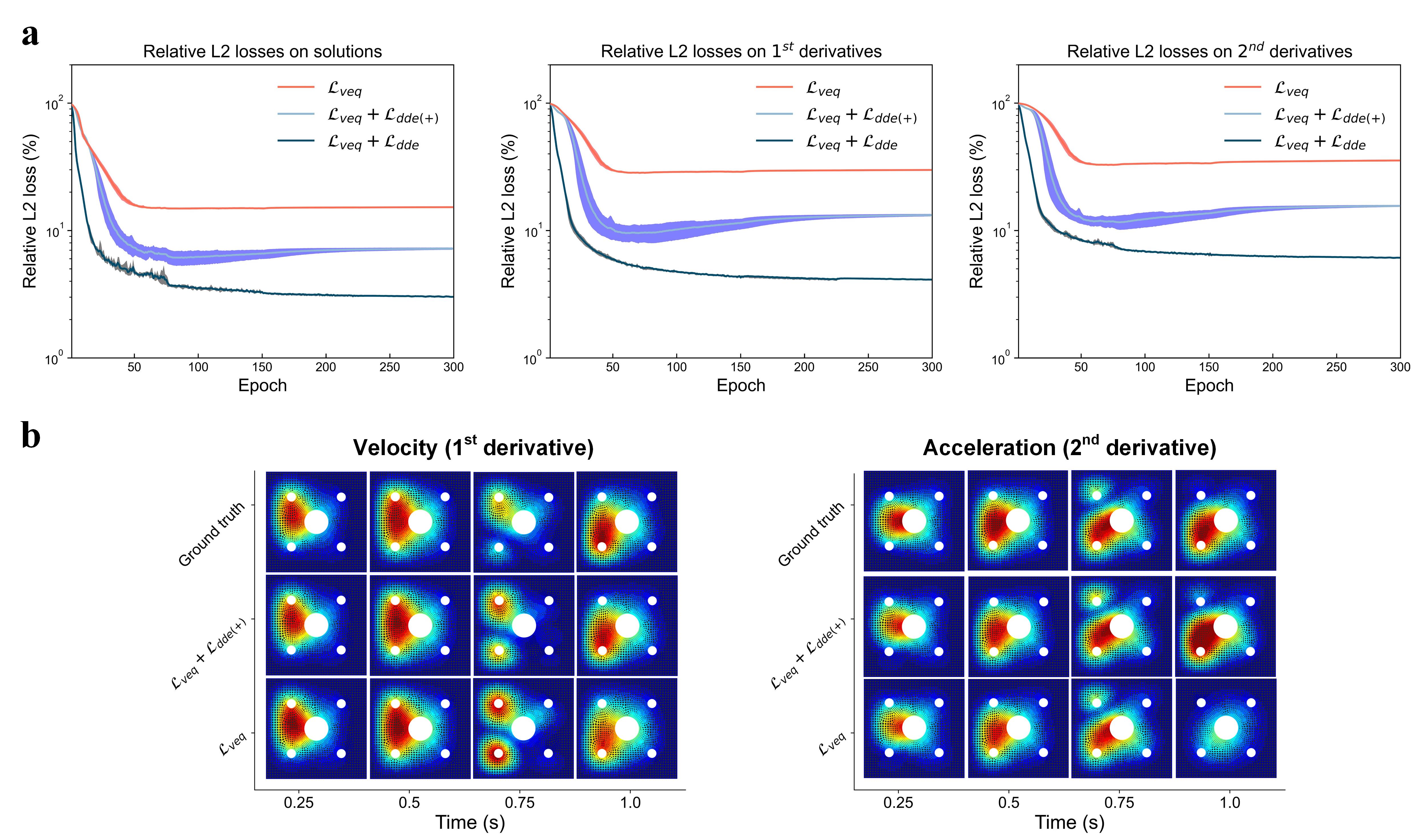}
\caption{\textbf{Performance on the simple mechanical structure (trained without data).} \textbf{a,} $rLSE$ for solutions, $1^{st}$ derivatives, and $2^{nd}$ derivatives during training. The red case was trained with $\mathcal L_{veq}$ while the cyan case considered $\mathcal L_{dde}$ on the edges of the time domain (0.05 s). The dark blue case represents a control case trained with $\mathcal L_{dde}$ over the entire time domain. \textbf{b,} Recovered velocity and acceleration fields for different cases.}
\label{Non-data_Results}
\end{figure*}

In theory, PINO-CDE can be trained without data, meaning only $\mathcal L_{eq}$ and $\mathcal L_{veq}$ are used. In addition, using $\mathcal L_{dde(+)}$ does not violate this concept, because the cost of solving the solution on the time domain boundaries is usually low in engineering simulations. We tested this idea on a simple mechanical system, as shown in Fig.~\ref{Non-data_Schematic}. Stochastic vibration loads were applied to three steel marbles, and fixed constraints were imposed on the perimeter of the steel plate at the bottom. Currently, the results obtained from the framework trained with only $\mathcal L_{veq}$ are unsatisfactory. However, $rLSE$ for the solutions can be reduced to less than 6$\%$ when boundary constraints $\mathcal L_{dde(+)}$ are provided, as shown in Fig.~\ref{Non-data_Results}. We envision that general non-data training strategies can be realized in future work by taking light from the existing research on PINN.

\begin{figure*}
\centering
\includegraphics[width=.9\linewidth]{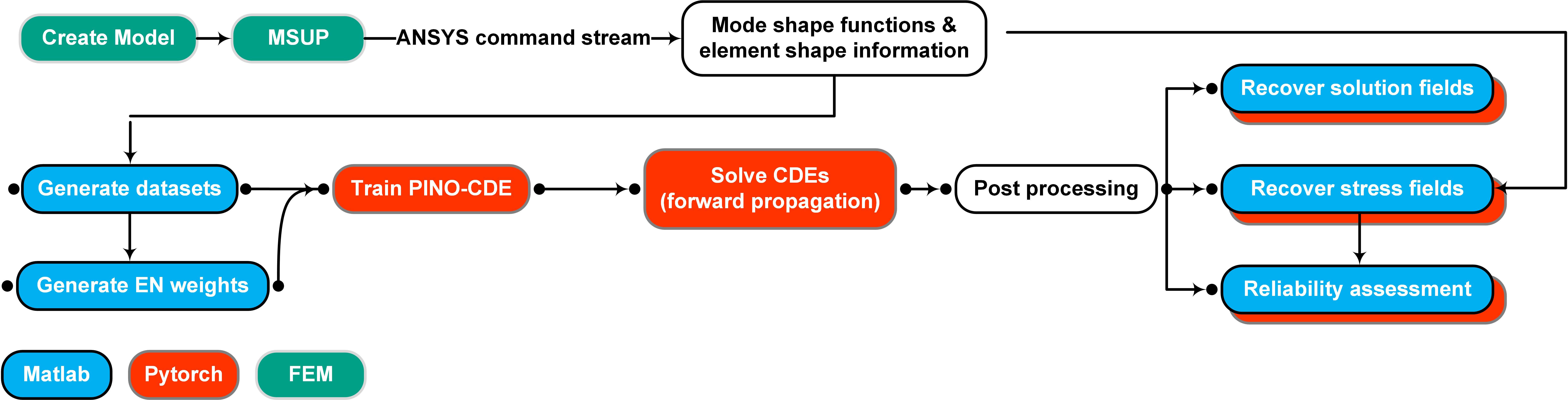}
\caption{\textbf{Instructive flowchart for the package.}}
\label{Reproduction_Guide}
\end{figure*}

\section {Reproduction Guide}
All codes for the reproduction of this paper have been uploaded to \href{https://github.com/Hedawl/PINO-CDE}{\textbf{Github}}. Datasets will be shared by request due to Github's file size limitation. In general, using the package includes several steps, as shown in Fig.~\ref{Reproduction_Guide}. Firstly, the numerical integration implemented by Matlab codes is used to generate the datasets. The training of PINO-CDE is done in the python environment using Pytorch. There are generally three goals for postprocessing: recovering solution (and derivatives) fields, recovering stress fields, and performing reliability assessments (uncertainty propagation). All three goals are implemented in the Matlab environment. Moreover, corresponding visualization codes are also provided, allowing users to observe the recovered time-variant solutions and probabilistic fields.\par
It should be noted that achieving these goals requires using structural mode shape functions and shape information for all finite elements of the structure. In this paper, we mesh the FEM model in ABAQUS and then import it to ANSYS through Hypermesh to obtain the mode shape functions and element shape information. The original FEM model files themselves are not provided due to size limitation. However, the processed mode shape functions and element shape information required to reproduce all the examples in this paper are provided directly. Moreover, an ANSYS command stream for users to export these data for their own models is given. More detailed instructions can be found on the project webpage: \href{https://github.com/WenHaoDing/PINO-CDE}{https://github.com/WenHaoDing/PINO-CDE}.




\clearpage
\bibliographystyle{elsarticle-num}
\bibliography{Bibliography}







\end{document}